\documentclass{svproc}

\usepackage{graphicx} 
\usepackage{epstopdf}
\DeclareGraphicsExtensions{.eps}

\usepackage{multirow}
\usepackage{makecell}
\usepackage{tabularx}
\usepackage{algorithmic}
\usepackage{algorithm}
\usepackage{amsmath} 
\usepackage{amssymb}
\usepackage{amsxtra}
\usepackage{url} 
\usepackage{color} 
\usepackage{bm}
\usepackage{placeins}
\usepackage[caption=false]{subfig}
\usepackage{xspace}
\usepackage{rotating}
\usepackage{siunitx}
\usepackage{textcomp}%
 
\newcommand{\wrt}{\text{w.r.t.}\xspace}

\newcommand{\myvector}[1]{\bm{#1}}
\newcommand{\myvec}[1]{\myvector{#1}}

\newcommand{\R}[1]{\mathbb{R}^{#1}}

\newcommand{\best}[1]{\textbf{#1}}
\newcommand{\algorithmicinput}{\textbf{Input:}}
\newcommand{\algorithmicoutput}{\textbf{Output:}}
\newcommand{\algorithmicinoutput}{\textbf{Input / Output:}}

\newcommand{\INPUT}{\item[\algorithmicinput]}
\newcommand{\OUTPUT}{\item[\algorithmicoutput]} 
\newcommand{\INOUTPUT}{\item[\algorithmicinoutput]}

\newcommand{\dur}[1]{\SI{#1}{\second}}

\newcommand{\excise}[1]{}

\newcommand{\rev}[2]{\textcolor{black}{#2}}

\newif\ifremark
\long\def\remark#1{
  \ifremark%
  \begingroup%
  \dimen0=\textwidth
  \advance\dimen0 by -1in%
  \setbox0=\hbox{\parbox[b]{\dimen0}{\protect\em #1}}
  \dimen1=\ht0\advance\dimen1 by 2pt%
  \dimen2=\dp0\advance\dimen2 by 2pt%
  \vskip 0.25pt%
  \hbox to \textwidth{%
    \vrule height\dimen1 width 3pt depth\dimen2%
    \hss\copy0\hss%
    \vrule height\dimen1 width 3pt depth\dimen2%
  }%
  \endgroup%
  \fi}

\newcommand{\ui}[1]{a}

\newcommand{\cspace}{\mathcal{C}}
\newcommand{\cfree}{\mathcal{C}_f}
\newcommand{\ecspace}{\mathcal{C}_e}
\newcommand{\ecfree}{\mathcal{C}_{ef}}
\newcommand{\workspace}{\mathcal{W}}
\newcommand{\cworkspace}{\mathcal{C}_w}
\newcommand{\dofw}{d_w}
\newcommand{\dofr}{d_r}
\newcommand{\dofe}{d_e}
\newcommand{\posew}{\textbf{w}}
\newcommand{\pose}{\textbf{p}}
\newcommand{\poser}{\textbf{p}_r}
\newcommand{\startr}{\textbf{p}_{rs}}
\newcommand{\goalr}{\textbf{p}_{rg}}
\newcommand{\epath}{\mathcal{P}}
\newcommand{\dataset}{\mathcal{D}}
\newcommand{\cnn}{C_{\myvec{\theta}}}
\newcommand{\tcnn}{C_{\hat{\myvec{\theta}}}}
\newcommand{\tbcnn}{\textbf{C}_{\hat{\myvec{\theta}}}}

\newcommand{\posei}{\textbf{p}^{(i)}}
\newcommand{\poseiminus}{\textbf{p}^{(i-1)}}
\newcommand{\poseione}{\textbf{p}^{(i+1)}}
\newcommand{\delp}{\Delta \textbf{p}^{(i)}}
\newcommand{\gp}{\textbf{g}^{(i)}}
\newcommand{\ggp}{\textbf{q}^{(i)}}

\newcommand{\nextra}{n_{\text{ex}}}

\newcommand{\tcnrrt}{t_{\text{cnrrt}}}
\newcommand{\trrt}{t_{\text{rrt}}}
\newcommand{\numqueries}{N}
\newcommand{\numcd}{n_{\text{checks}}}
\newcommand{\tgjk}{t_{\text{check}}}

\newcommand{\datasetsize}{\| \dataset\|}
\newcommand{\tinvest}{t_{\text{invest}}}
\newcommand{\toffline}{t_{\text{offline}}}

\newcommand{\ffp}{f_{\text{fp}}}
\newcommand{\ffn}{f_{\text{fn}}}
\newcommand{\fspeedup}{f_{\text{batch}}}
\newcommand{\ftraining}{f_{\text{train}}}
\newcommand{\pathlength}{f_{\text{pathlen}}}

\usepackage{subfig, graphicx, algorithm, algorithmic, amsmath, amssymb, wrapfig, afterpage, makecell, url}

\let\llncssubparagraph\subparagraph
\let\subparagraph\paragraph
\usepackage{titlesec}
\titlespacing*{\section}{0pt}{0.7\baselineskip}{0.7\baselineskip}
\titlespacing*{\subsection}{0pt}{0.7\baselineskip}{0.7\baselineskip}
\let\subparagraph\llncssubparagraph

\begin{document}
\mainmatter
\title{Neural Collision Clearance Estimator for Batched Motion Planning}
\titlerunning{Neural Collision Clearance}  
%
\author{J. Chase Kew\inst{1} \and Brian Ichter\inst{1} \and
Maryam Bandari\inst{2} \and Tsang-Wei Edward Lee\inst{1} \and Aleksandra Faust\inst{1}}
\authorrunning{J. Chase Kew et al.} 
\tocauthor{J. Chase Kew, Brian Ichter, Maryam Bandari, Tsang-Wei Edward Lee, Aleksandra Faust}
\institute{Robotics at Google, Mountain View, CA, USA,
\email{\{jkew,ichter,tsangwei,faust\}@google.com}
\and
X, Mountain View, CA, USA,
\email{maryamb@x.team}
}

\maketitle              

\begin{abstract}
We present a neural network collision checking heuristic, ClearanceNet, and a planning algorithm, CN-RRT.
ClearanceNet learns to predict separation distance (minimum distance between robot and workspace) with respect to a workspace.
CN-RRT then efficiently computes a motion plan by leveraging three key features of ClearanceNet.
First, CN-RRT explores the space by expanding multiple nodes at the same time, processing batches of thousands of collision checks. Second, CN-RRT adaptively relaxes its clearance requirements for more difficult problems. Third, to repair errors, CN-RRT shifts its nodes in the direction of ClearanceNet's gradient and repairs any residual errors with a traditional RRT, thus maintaining theoretical probabilistic completeness guarantees. In configuration spaces with up to 30 degrees of freedom, ClearanceNet achieves 845x speedup over traditional collision detection methods, while CN-RRT accelerates motion planning by up to 42\% over a baseline and finds paths up to 36\% more efficient. Experiments on an 11 degree of freedom robot in a cluttered environment confirm the method's feasibility on real robots. 

\end{abstract}

\vspace{-1.5\baselineskip}
\newcommand{\pichgt}{4.5cm}
\begin{figure}[t]
\centering
\begin{tabular}{cc}
\subfloat[Start pose]
    {\includegraphics[height=\pichgt,keepaspectratio=true]{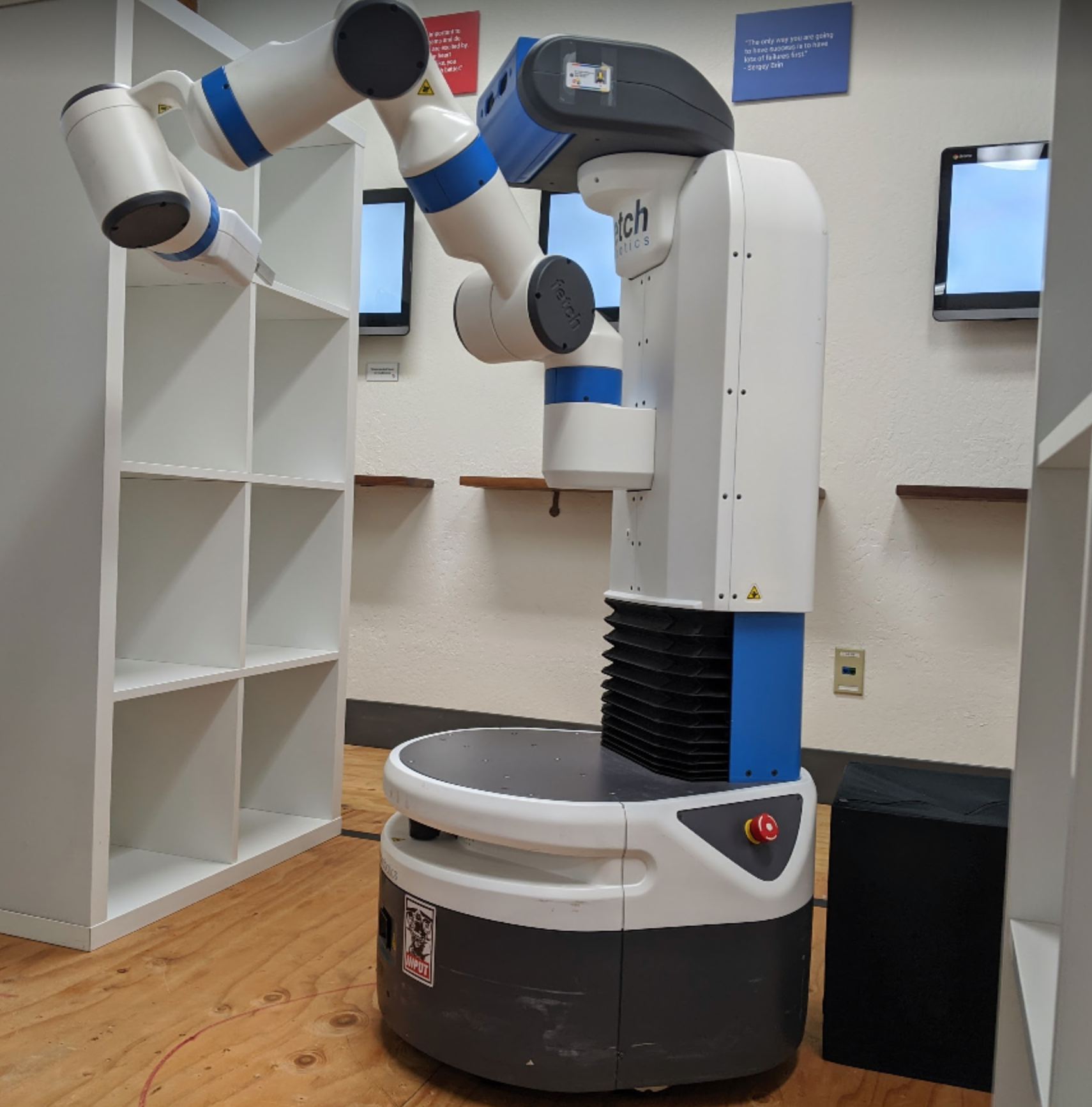}}&
\subfloat[End pose]
    {\includegraphics[height=\pichgt,keepaspectratio=true]{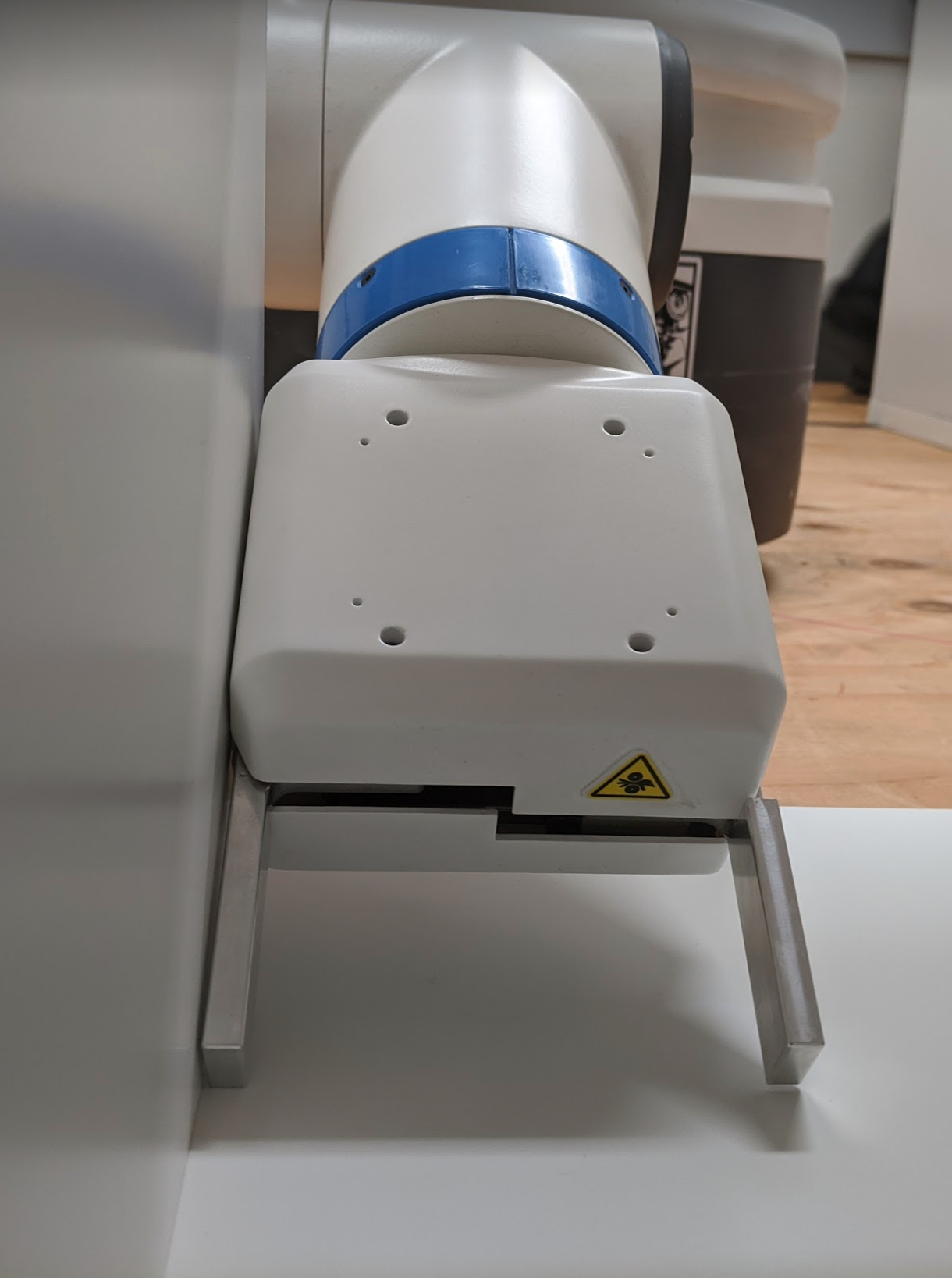}}
\end{tabular}
\vspace{-0.75\baselineskip}
	\caption{\footnotesize Real-world validation with Fetch robot reaching into a shelf.
	\label{fig:fetch_real}}
\vspace{-1.5\baselineskip}
\end{figure}

\section{Introduction}
\begin{figure*}[tb]

  \resizebox{1.0\textwidth}{!}{
    \subfloat[Block, 14 DoF]
    {\includegraphics[height=5.0cm,keepaspectratio=true]{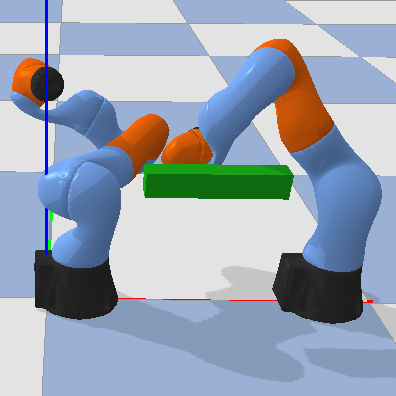}\label{fig:two_arms_block}} 
    \quad
    \subfloat[R2D2, 14 DoF]
    {\includegraphics[height=5.0cm,keepaspectratio=true]{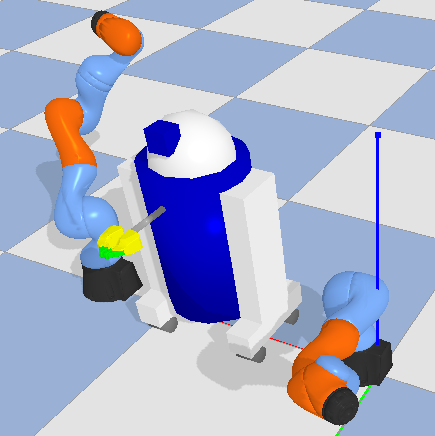}\label{fig:two_arms_r2d2}} 
    \quad
    \subfloat[Ducky, 7 DoF]
    {\includegraphics[height=5.0cm,keepaspectratio=true]{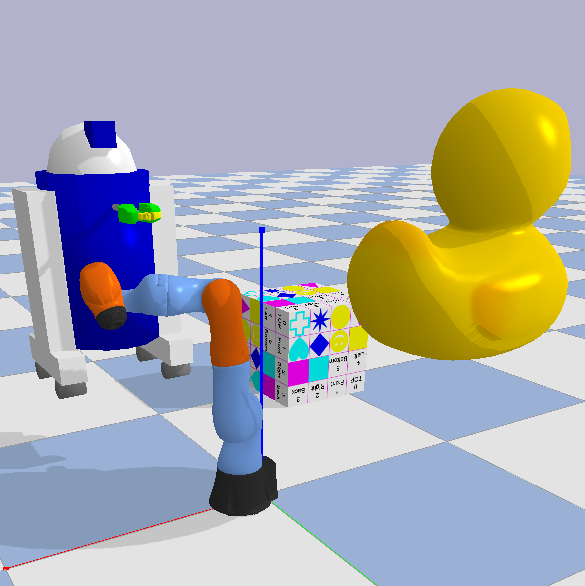}\label{fig:ducky}} 
    \quad
    \subfloat[Mobile, 30 DoF]
    {\includegraphics[height=5.0cm,keepaspectratio=true]{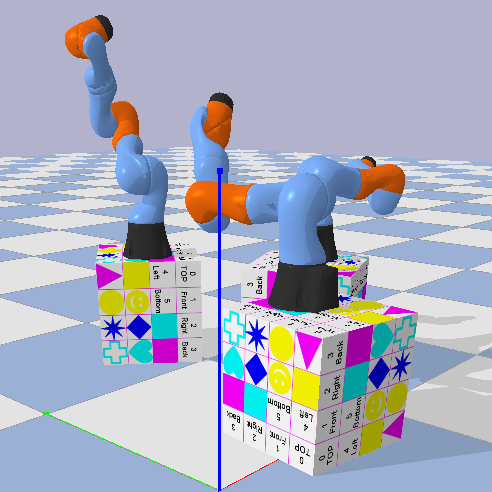}\label{fig:three_arms_mobile}} 
    \quad
    \subfloat[Fetch, 11 DoF]
    {\includegraphics[height=5.0cm,keepaspectratio=true]{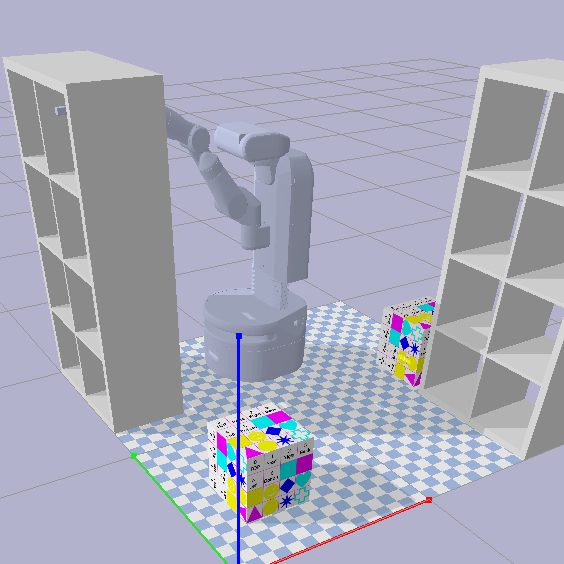}\label{fig:fetch}}  
}
	\caption{\footnotesize Environments: 
	a), b) Static workspace with fixed robot bases. 
	c) Parameterized workspace - robot is fixed while obstacle placement varies. 
	d) Mobile manipulators. Model trained for two mobile manipulators and used to plan for three without additional training.
	e) Fetch robot with 7 DoF arm, torso lift and base position. Each set of shelves has one DoF and cubes have 2 DoFs each.
	\label{fig:envs}}
\vspace{-1\baselineskip}
\end{figure*}

\rev{E.1 3.1}{Modern robots work in complex workspaces such as factory lines and warehouses. 
They shelve inventory (Fig. \ref{fig:fetch_real}), inspect parts (Fig. \ref{fig:two_arms_r2d2}), and perform collaborative tasks (Fig. \ref{fig:three_arms_mobile}). 
Within its lifetime a robot can complete hundreds of thousands of tasks. 
In factories and warehouses the workspaces change in structured ways: objects are displaced as the robot stocks the shelves, and free space depends on other robots' poses. 
However, classic motion planning takes little advantage of this structure in changing environments and typically plans every path from scratch. 
In this work, we present a learning-based method that leverages prior experience to quickly propose paths and pass them to a computationally expensive geometry-based planner for validation and repair.
} 

Motion planning is the problem of finding a feasible path between start and goal poses in the robot's workspace. 
Often, motion planning problems are solved in configuration space ($\cspace$), the set of possible robot poses \cite{Lavalle06book} within a static workspace.
As the dimensionality of $\cspace$ increases, exactly solving the planning problem quickly becomes intractable.
Sampling-based planning addresses this issue by forming an approximate, implicit representation of  $\cspace$ through a set of probing random samples, connected locally by querying a collision checker.

Classical approaches to collision checking rely on computing complicated geometries. 
Though there exists a rich literature on efficient computation \cite{Lin91afast,dinesh-collision}, accelerating collision checking through parallelization, broad and narrow search phases and other optimizations, the exact geometric collision check remains a computationally expensive operation.
Researchers have attempted to reduce the number of required collision checks by certifying regions to be collision free~\cite{bialkowski2016efficient} or by using probabilistic collision checks~\cite{kumar2019cspacebelief} to provide a quick, approximate belief about which regions are in collision. Regardless, collision checking remains the primary bottleneck in sampling-based motion planning, consuming up to $90\%$ of total computation time~\cite{elbanhawi2014review}. 

We address the collision checking bottleneck and accelerate sampling-based planning with three key insights.
First, separation distance is a function of only the poses of all mobile bodies in an environment. 
This leads to the idea of learning to predict separation distance (clearance) from examples,
without the robot explicitly knowing either its own geometry or the obstacles' during planning.
This works because deep neural networks are universal approximators that with enough data can learn any continuous function
and because clearance is a Lipschitz continuous function \cite{Lavalle06book}.
We therefore train a deep neural network, ClearanceNet, to predict collision clearance for robots and workspaces with many degrees of freedom (DoFs), using geometry for training but not for inference.
The second insight is that both sampling-based motion planning and neural network inference are embarrassingly parallelizable \cite{nancy-parallel,bialkowski2011massively}. Leveraging neural networks' efficiency at processing large batches, we collision check entire edges together and expand from many RRT nodes in parallel.
This batch processing leads to a heuristic collision check two orders of magnitude faster than optimized geometric methods on a single GPU.
The third insight is that often it is easier to verify and repair a partially-correct trajectory than to plan from scratch. This is relevant because, although fast, the learned collision heuristic is approximate and makes mistakes. Thus, building on~\cite{das2020fastron}, we verify and repair motion plans using the gradient of the neural network to shift poses towards higher clearance, then repair remaining misclassified poses with traditional RRT \cite{rrts}.

Using these insights, this paper makes three contributions. 
1) \textit{ClearanceNet}, a neural network that estimates minimum separation distance, conditioned on joint robot-workspace pose. ClearanceNet is trained using large-scale hyperparameter tuning that simultaneously searches for the appropriate training hyperparameters and trains the clearance estimator. We also provide the training hyperparameters, which were consistent across environments and robots.
2) \textit{CN-RRT}, an adaptive and batched extension of the Fastron-RRT algorithm \cite{das2020fastron}, which leverages the efficiency of neural network inference when processing large batches of data (up to 60 edges and 5,000 collision checks). CN-RRT adaptively reduces the difficulty of the motion planning problem and includes a repair step that attempts to move the path away from obstacles using ClearanceNet's gradient.
3) A configurable workspace, extended configuration space and in-depth analysis of CN-RRT trade offs.

ClearanceNet and CN-RRT are evaluated on five environments (Fig. \ref{fig:envs}) with convex and concave obstacles. Robots range from 7 to 30 DoFs, including multiple arms with fixed bases, mobile manipulators, and a full bodied Fetch robot. 
We compare the presented method with two baselines: a traditional C++ optimized 
geometric collision checker (Gilbert-Johnson-Keerthi or GJK~\cite{gjk1988}) implemented in PyBullet \cite{pybullet}, and Fastron \cite{das2020fastron}\rev{R1.4}{, a learned collision checker with a kernel perceptron}.  ClearanceNet performs collision checking up to 845x faster than GJK in sufficiently large batches, and achieves accuracy of 91-96\%. CN-RRT both completes planning faster (up to 42\%) and produces up to 36\% shorter paths. 

\section{Related Work}
\label{sec:rel-work}

Several previous works used machine learning to offset the cost of collision checking.
Some methods learn heuristics for promising paths \cite{chiang2018fast}, learn a distribution of promising regions \cite{ichter2018learning}, or evaluate only the most promising edges \cite{choudhury2018data} \rev{R2.5}{based on predicted energy costs \cite{yavari2019lazy}}.
Our approach, which learns to accelerate the actual collision checking, can be used in conjunction with these approaches.

Other works use learning to accelerate collision checking itself by approximating $\cspace$ with Gaussian mixture models \cite{huh2016learning}, or by classifying collisions based on nearby collision checks \cite{kumar2019cspacebelief,pan2016fast}. Another work trains a support vector machine offline and actively adapts it online to predict the penetration depth of rigid bodies \cite{pan2013efficient}.
Our method, however, focuses specifically on learning to classify collisions in the configuration space and ultimately certifies path validity.

In \cite{chinwe} a contractive autoencoder and multi-layer perceptron predict collisions from latent space in a probabilistic roadmap. Our work differs in that: \rev{R3.4}{1) batching speeds up planning, 2) a repair step guarantees correctness and 3) inputs are configuration points instead of occupancy grids, allowing us to expand the workspace volume without changing the neural network architecture and to support obstacles with arbitrary geometry, beyond axis-aligned boxes.}
The work most similar to ours is Fastron \cite{das2020fastron}, which uses a kernel perceptron and active learning to form a belief model of $\cspace$. Because Fastron is not based on a neural network, it cannot take advantage of the batch processing speedup. Additionally, our method generalizes to object repositioning in the workspace without further training. CN-RRT takes Fastron-RRT \cite{das2020fastron} as a foundation and adds whole-edge collision checks, edge building parallelization, adaptive thresholding, and a flexible repair step to guarantee probabilistic completeness. 
New simulation environments are emerging for path planning that allow users to programmatically create new workspaces by randomly placing a fixed number of objects within a static environment \cite{interactivegibson}. Here we consider a similar setting, where the workspace consists of fixed obstacles and obstacles that may change positions between two motion planning problems.

\section{Problem Statement}
\label{sec:ps}
In this paper we use relaxed assumptions about robot workspace and configuration space. 
To that end, this section introduces key definitions for workspace configuration space, extended configuration space and our problem definition.

\begin{definition}
\textbf{Workspace configuration space.} $\cworkspace \subset \R{\dofw}$ is the set of all possible workspace configurations for $\dofw \geq 0.$ Vector $\posew \in \cworkspace$ is a \textbf{workspace configuration} and scalar $\dofw$ is the \textbf{workspace degrees of freedom}. We denote with $\workspace(\posew)$ a workspace with configuration $\posew$.
\end{definition}
Intuitively, a workspace configuration space corresponds to a set of related workspaces where large portions are fixed, but there are several objects that can move between planning problems. In a typical room, for example, the floor, ceiling, tables and cabinets remain static, while the locations of chairs may change. In the special case $\dofw = 0,$ $\cworkspace = \varnothing,$ and there is only one possible workspace corresponding to the classic static workspace $\workspace.$

\begin{definition}
\textbf{Extended configuration space.} $\ecspace$ is the set of all possible robot and workspace configurations, $\ecspace = \cspace \times \cworkspace \subset \R{\dofe},\,\dofe = \dofr + \dofw,$ where $\cspace \subset \R{\dofr}$ is the set of all possible robot poses $\poser.$ A configuration space for a fixed workspace configuration $\posew$ is denoted by $\ecspace(\posew)$.
\end{definition}
The extended configuration space, a configuration space generalization, contains $\dofr$ robot controllable DoFs, and $\dofw$  DoFs not controllable by the robot. Robot pose $\poser$ is a joint configuration for all robots in the workspace for which we are finding a motion plan, while the workspace may contain objects that change positions between motion planning instances. 
In the special case $\dofw = 0$ all DoFs of the extended configuration space are controllable by the robot, and $\ecspace = \cspace.$

\begin{definition}
\textbf{Collision checking in $\ecspace.$} A configuration $\pose = (\poser,\posew)$ is valid iff robot configuration $\poser$ is valid
in workspace $\workspace(\posew).$ \textbf{Extended free configuration space}, $\ecfree \subset \ecspace$ is the set of all valid
configurations $\ecfree = \{\pose = (\poser,\posew) |\,isValid(\poser, \workspace(\posew)),\, \poser \in \cspace,\, \posew \in \cworkspace \}.$ The free configuration space for a fixed $\posew$ is denoted by $\cfree(\posew)$.
\end{definition}
In other words, a configuration is valid in joint robot-workspace configuration space if the robot pose is valid in this particular instance of the workspace. 

\begin{definition}
\textbf{Path planning in extended configuration space.} Given extended configuration space $\ecspace,$ workspace configuration $\posew,$ start and end robot poses, $\startr, \goalr$, find a valid path $\epath = [\pose_{r0},\cdots, \pose_{rn}],$ such that the
path starts in the start robot pose, ends in the end robot pose, and each point in the path is valid: $\pose_{r0} = \startr,\, \pose_{rn} = \goalr, \pose_{ri} \in \cfree(\posew)$ for $i = 1,\cdots,n.$ 
\end{definition}

\begin{definition}
\textbf{Data batch.} Let $\pose \in \R{\dofe}$ be a vector, $\mathcal{B} = [ \pose_1, \, \cdots,\,\pose_n]^\intercal$ be a $\dofe \times n$ matrix of batched data, and $\tcnn:\R{dofs} \rightarrow \R{}$ be a neural network with $\dofe$ inputs and one output. Batched inference $\tbcnn$ maps a batched data matrix $\mathcal{B}$ into an n-dimensional vector.
\end{definition}
Deep neural networks are optimized to efficiently process large batches of data using low-level parallelization of matrix multiplication. Leveraging this property of neural networks is key in our algorithm's multi-directional RRT growth.

\section{Methodology}
\label{sec:methods}

We introduce ClearanceNet, CN: $\ecspace \rightarrow \R{}$, a learned clearance estimator, and its use within a sampling-based motion planning algorithm, CN-RRT.
ClearanceNet is a neural network that takes a configuration (e.g. a robot pose and workspace configuration) and estimates the minimum clearance between robot and workspace.
CN-RRT, a batched variant of Rapidly-exploring Random Trees (RRTs), then explores $\cspace$ under the supervision of ClearanceNet and certifies (and possibly repairs) the final motion plan with exact collision checks.

\subsection{ClearanceNet} 

\textbf{Data Collection:} 
The data collection process samples a robot pose $\poser$ and workspace configuration $\posew$ from the extended configuration space. It sets the workspace to $\workspace(\posew)$ and robot to $\poser$ and computes the clearance $d,$ the minimum distance between the robot and the workspace or itself, using a geometric method \cite{gjk1988}. Positive values indicate the configuration point is free of collision, i.e. $\poser \in \ecfree(\posew),$ while negative values indicate penetration depth. The triples are then added to a dataset:
$\dataset = \{(\poser,\,\posew,\, d)\, |\, (\poser,\posew) \sim \ecspace, d \in \R{} \}.$
When the dataset is large enough, it is partitioned into training and evaluation sets.

\textbf{ClearanceNet Training:} We train ClearanceNet from the collected dataset $\dataset$. Let $\cnn: \ecspace \rightarrow \R{}$ be a neural network parameterized with weights $\myvec{\theta}$ that takes a robot and a workspace configuration point $(\poser,\, \posew)  \in \ecspace$ as input and predicts the minimum clearance $d$ in the workspace $\workspace(\posew).$ The neural network consists of two fully connected layers, each followed by a dropout layer, and an output layer that predicts clearance. The training minimizes the mean squared error,
$\hat{\myvec{\theta}} = \arg \min_{\myvec{\theta}} N^{-1}\sum_{i=1}^{N} \|d^{(i)} - \cnn(\poser^{(i)}, \posew^{(i)})\|^2.$
Finding the right neural network size, dropout rates, learning rate, and minibatch size is not trivial and often time consuming. We automate the search using a large-scale hyperparameter optimization according to Gaussian Process Bandits \cite{gp-bandits}. This process trains a population of neural nets with different parameters and observes their performance, then selects hyperparameters for the next generation until the hyperparameters are tuned. A trained ClearanceNet is denoted by $\tcnn$.

\textbf{Collision classifier definition:} 
After the clearance estimator $\tcnn$ is trained, we make a collision classifier by selecting a clearance threshold $d^*$.
A pose $\poser$ is classified as in collision if $\cnn(\poser, \posew) < d^*$, i.e. the predicted clearance is less than the threshold.
We can vary $d^*$ to obtain a stricter or more permissive classifier.

\begin{algorithm}[tb]
    \caption{CN-RRT}\label{alg:parallelConnect}
\begin{algorithmic}[1]
\INPUT  Initial and goal robot poses: $\startr, \goalr$ and a workspace configuration: $\posew$.
\INPUT $\tbcnn:$ Trained clearance estimator.
\INPUT $\myvec{d}^*, \myvec{\Delta t}$: Monotonically decreasing vectors of CN thresholds and times to switch.
\INPUT $N_{\text{ext}}$: Number of simultaneous extensions. 
\INPUT $t_{\text{max}}$: Timeout. 
\OUTPUT $\epath$: Sequence of robot poses from $\startr$ to $\goalr$.
\STATE Initialize tree $\mathcal{T}$ with $\startr$ and current threshold index $j=1.$
\WHILE {$\goalr$ not reached and $t_\text{elapsed} < t_{\text{max}}$}
  \STATE $P_\text{sample} \leftarrow [\poser^{(1,s)},\cdots,\poser^{(N_{\text{ext}},s)} | \poser^{(i,s)} \sim \ecspace(\posew)].$ Sample $N_{\text{ext}}$ points from $\ecspace(\posew).$
  \STATE $P_\text{expand} \leftarrow [\poser^{(1,e)},\cdots,\poser^{(N_{\text{ext}},e)} | \poser^{(i,e)}:$ Nearest in $\mathcal{T}$ to $\poser^{(i,s)}].$ Nodes to expand.
  \STATE $P \leftarrow $ Discretized paths from $\poser^{(i,e)}$ to  $\poser^{(i,s)}$ for each $i = 1,\cdots, N_{\text{ext}}$
  \STATE Find clearances $\textbf{c} \leftarrow \tbcnn(\mathcal{B}),$ for the batch of points, $\mathcal{B} = \{(\poser, \posew)\ |\ \forall \poser \in P\}.$
  \STATE Truncate each path at first clearance $< \myvec{d}^*[j]$
  \STATE Randomly sample poses from truncated paths and add them to tree $\mathcal{T}$.
  \IF {$t_\text{elapsed} > \myvec{\Delta t}[j]$}
    \STATE $j \leftarrow \min(j+1, \text{length}(\myvec{d}^*))$. Estimator $\tcnn$ less conservative.
  \ENDIF
\ENDWHILE
\STATE \rev{1.1, 3.3}{$\epath \leftarrow$ \textbf{if} ($\goalr \in \mathcal{T}$) \textbf{then} Extract path from tree $\mathcal{T}$ \textbf{else} $[\startr, \goalr].$}
\RETURN \rev{1.1, 3.3}{Algorithm \ref{alg:repair} ($\epath$) /* Validate, and repair if necessary, path $\epath$. */}
\end{algorithmic}
\end{algorithm}

\subsection{CN-RRT: Sampling-based planning with ClearanceNet}
\label{sec:CN-RRT}

ClearanceNet serves as a heuristic to select promising edges in a sampling-based planning setting.  
To best utilize it, we make three modifications to the Fastron-RRT algorithm \cite{das2020fastron}. 
First, taking advantage of neural networks' ability to efficiently evaluate large batches of data, we present an algorithm that evaluates several entire edges for collisions in one large batch. 
Second, we introduce adaptive thresholding to control how conservative the algorithm is. Lowering the threshold reduces false positives (free space predicted to be in collision) but increases false negatives.
Last, we add a gradient-based path repair, which is possible because ClearanceNet is differentiable in configuration space. Intuitively, during repair we move the path away from obstacles in the direction of the gradient.

\textbf{Batch Node Expansion:}
Algorithm \ref{alg:parallelConnect} outlines CN-RRT. The tree grows from the start configuration, expanding until it reaches either the goal or the time limit. During each iteration, CN-RRT selects a batch of configuration points (Line 3) and finds their nearest neighbors in the tree (Line 4). Next, it computes the edges between the random points and their neighbors using a local planner, \textit{without} checking for collisions (Line 5). Then ClearanceNet predicts clearances for the entire batch that contains configuration points for all edges (Line 6). 
Next, for each edge, CN-RRT finds the first node where the predicted clearance is smaller than the allowed threshold, i.e predicted collision (Line 7). The algorithm then adds several samples from the estimated collision-free portion of the edge (Line 8). Each sample is added to the tree, and all are connected from the expansion node (Line 8). This step ensures addition of nodes that are not near collision boundaries, although it does increase the size of the tree. 

\textbf{Clearance Threshold Adaptation:} CN-RRT has an adaptive clearance threshold $d^*$ to make the algorithm more flexible (note that $d^*$ controls the precision of the classifier). 
An overly conservative classifier will unnecessarily discard collision-free samples, but most samples classified as collision free will be truly collision free. 
Many easy motion planning problems (less cluttered workspace, start and end poses nearby, etc.) can be solved even with an overly conservative classifier, and these will require little or no repair, but other problems require less conservative solutions. Because we do not know ahead of time what threshold is appropriate for a given problem, we give CN-RRT a list of thresholds and time limits as input. If it finds no solution within a given time limit (Line 9), the algorithm decreases the threshold (Line 10) to relax the classifier. This step ensures that in subsequent iterations more nodes are added to the tree, although they may not be truly collision free. 

\textbf{Gradient-based Path Repair:}\label{sec:grad}
Because the learned collision checker is approximate, CN-RRT may need to resolve misclassifications.
It does so with a repair step that uses the gradient of ClearanceNet.
Recall that ClearanceNet is a differentiable scalar function of robot pose $\poser$ and workspace state $\posew$ that predicts a clearance, $\tcnn(\poser, \posew) \rightarrow d.$ This gradient is automatically computed to train the network.
This means that for a robot pose in collision, following the positive gradient of the clearance with respect to robot pose can increase the clearance; initially reducing the amount of collision and eventually finding a collision-free point. 
Let $\posei, \poseione \in \mathcal{P}$ be two consecutive configurations points in the proposed path. Let $\gp = \nabla \tcnn (\posei)$ be a gradient of the ClearanceNet \wrt the point, and $\delp = \poseione - \posei$ be a vector in the direction of the path. 
$\posei$ moves in the positive gradient direction perpendicular to $\delp$:
\begin{equation}
    \label{eq:update}
    \posei \leftarrow \posei + \alpha \ggp,\;\;\;\;\ggp = \gp - (\gp \cdot \delp \|\delp\|^{-1})
\end{equation}
where $\alpha$ is step size and $\ggp$ is the orthogonal projection of gradient $\gp$ onto the hyperplane orthogonal to the path vector $\delp.$

\begin{algorithm}[t]
    \caption{Repair}\label{alg:repair}
\begin{algorithmic}[1]
\INPUT $\tbcnn:$ Trained clearance estimator. $\alpha:$ step size. $\nextra:$ extra steps. $t_{\text{max}}$: Timeout. 
\INOUTPUT $\epath$: Sequence of poses from $(\startr, \posew)$ to $(\goalr, \posew).$
\OUTPUT \rev{1.1, 3.3}{Success: Boolean. Is the path certified valid?} 
\STATE $i \leftarrow 0,$ \rev{1.1, 3.3}{$n_{\text{shifts}} \leftarrow 0$}
\WHILE{$0 < i<\text{length}(\epath)$ and \rev{3.3}{$n_{\text{shifts}} < t_{\text{max}}$}}
  \STATE \textbf{if} \text{isValid}($\posei$) \textbf{then} $i \leftarrow i+1$; Continue.
  \WHILE{not \text{isValid}($\posei$) \rev{1.1, 3.3}{and $n_{\text{shifts}} < t_{\text{max}}$}}
    \STATE Shift $\posei$ toward safety by $\alpha$ according to Eq. \eqref{eq:update}. \rev{1.1, 3.3}{$n_{\text{shifts}} \leftarrow n_{\text{shifts}} + 1$}
  \ENDWHILE
  \STATE Shift $\posei$ another $\nextra$ steps toward safety. 
  \STATE Add unchecked interpolation points between new $\posei$ and $\poseiminus$, $\poseione.$  
  \STATE $i \leftarrow i$ Now refers to new next point after $\poseiminus.$
\ENDWHILE
\STATE \rev{1.1, 3.3}{\textbf{if} {$i == \text{length}(\epath)$} \textbf{return} True, $\epath$ {/* Whole path is certified valid. */}}
\FOR {\rev{1.1, 3.3}{$j=i-1; j > 0; j \leftarrow j-1$ /* Back out and repair with RRT.*/}} 
\STATE \rev{1.1, 3.3}{\textbf{if} Fastron repair step ~\cite{das2020fastron} on $\epath(j, \epath(\text{length}))$ is successful \textbf{then} return True, $\epath$.}
\ENDFOR
\RETURN \rev{1.1, 3.3}{False, $\epath$ \rev{1.1, 3.3}{/* RRT could not find a valid path */}}
\end{algorithmic}
\end{algorithm}

\begin{figure*}[h]
  \resizebox{1.0\textwidth}{!}{
    \subfloat[\scriptsize Find path.]
    {\includegraphics[height=2.5cm,keepaspectratio=true]{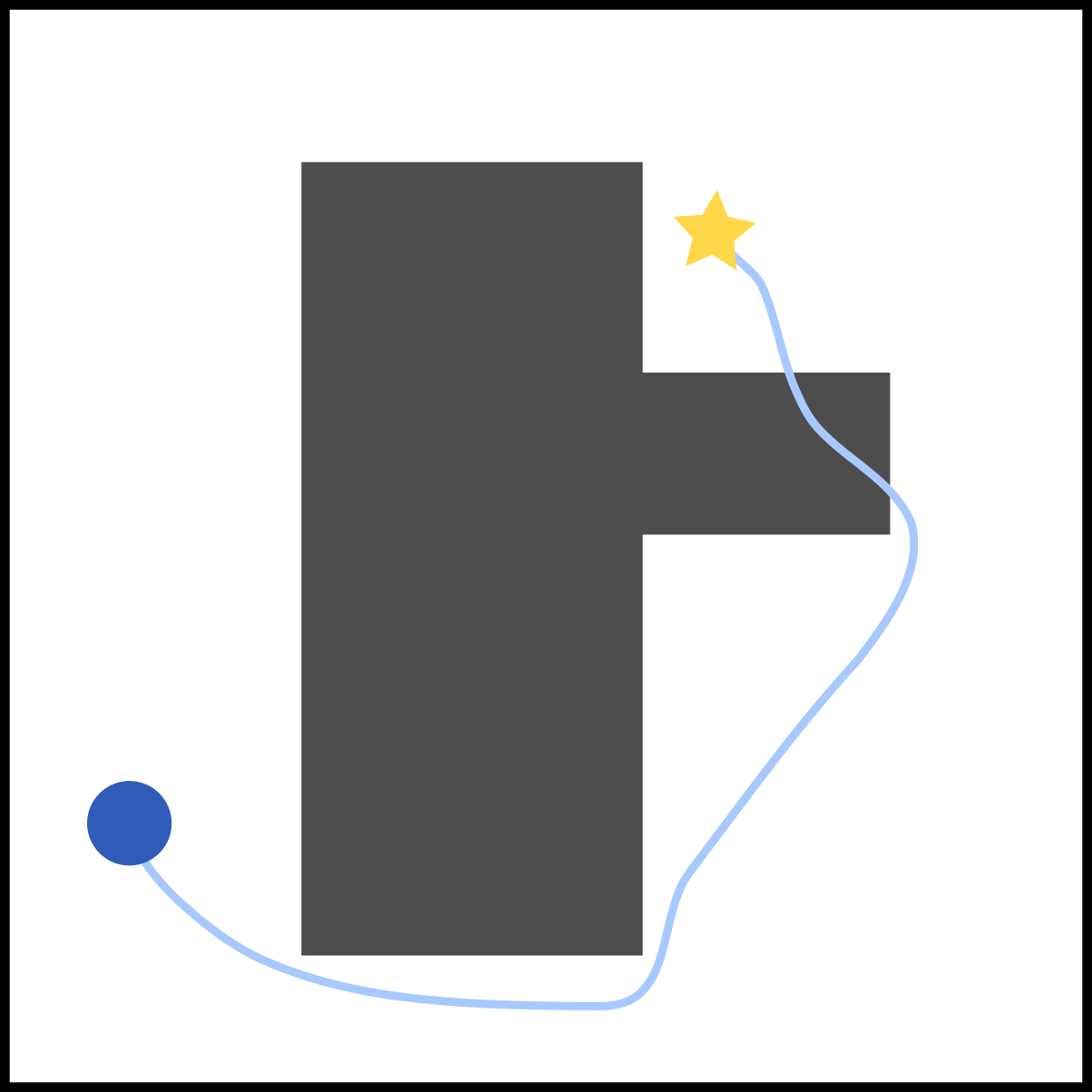}} 
    \quad
	\subfloat[\scriptsize Check path.]
	{\includegraphics[height=2.5cm,keepaspectratio=true]{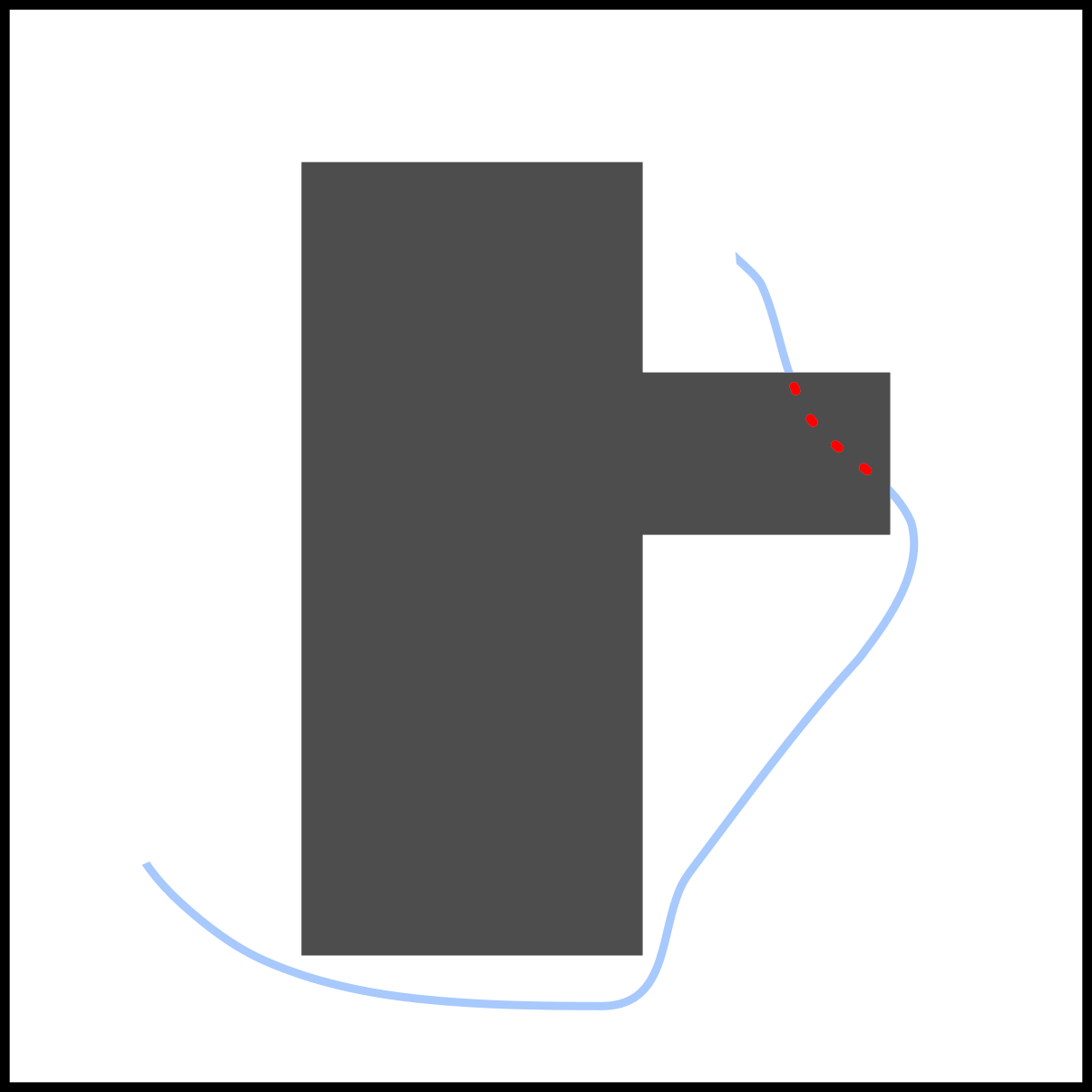}} 
    \quad
    \subfloat[\scriptsize Define new problem.]
	{\includegraphics[height=2.5cm,keepaspectratio=true]{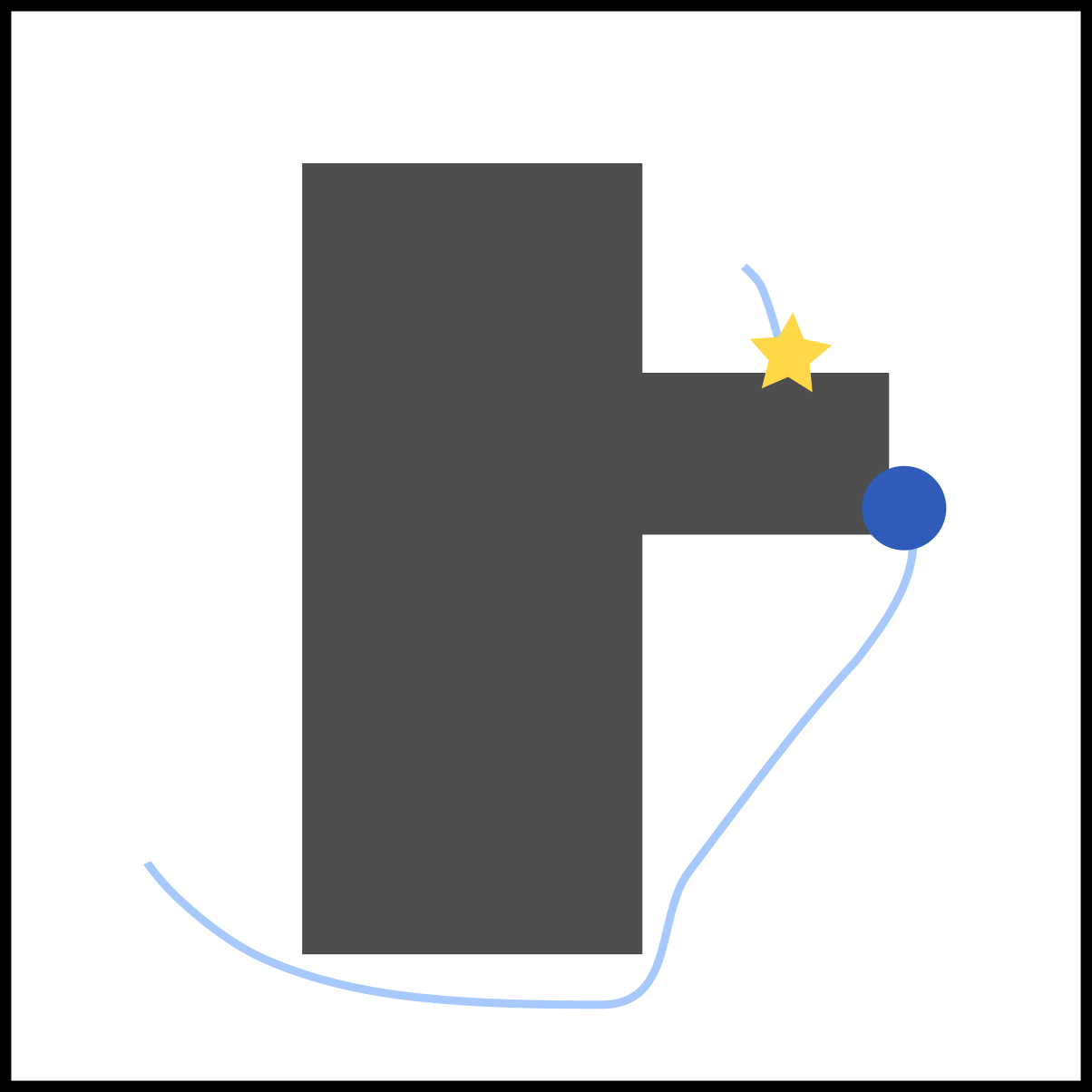}} 
    \quad
    \subfloat[\scriptsize Combine results.]
    {\includegraphics[height=2.5cm,keepaspectratio=true]{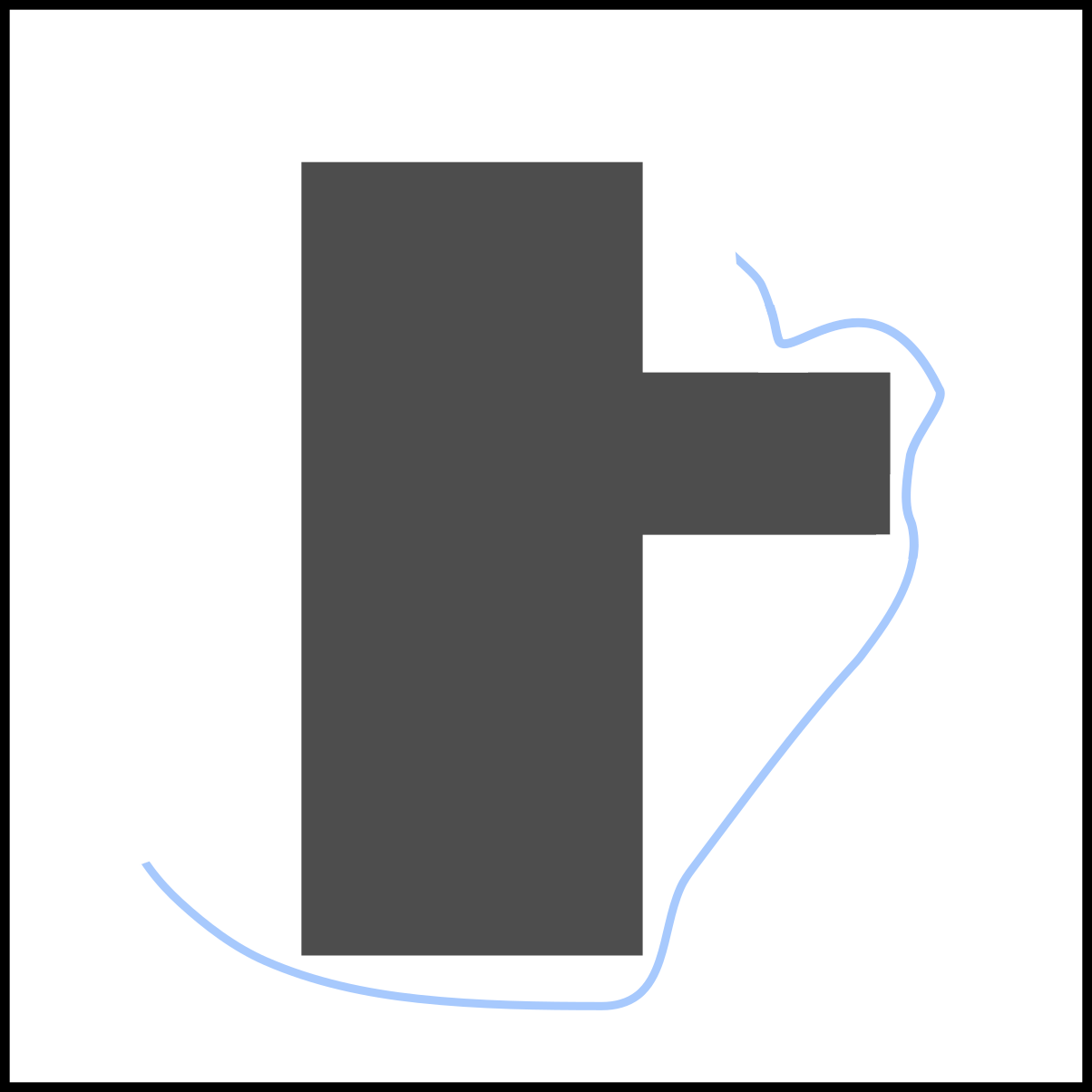}} 
}   
	\caption{\footnotesize The steps of the Fastron-RRT repair algorithm. (a) We use Fastron-RRT to find a path from the blue circle to the yellow star. Because Fastron is a heuristic, the path may have collisions. (b) We check the path point-by-point using a geometry-based collision checker. The red points are in collision, so we excise them from the path. (c) We define a new planning problem from the blue circle to the yellow star, to circumvent the in-collision section. (d) We solve the new planning problem using GJK-RRT and insert the result into the path.}
	\label{fig:fastron-repair}
\end{figure*}

Algorithm \ref{alg:repair} describes the repair procedure. First, we check all points in the path with classic geometry-based collision checking (Line 3). \rev{1.1, 3.3}{If all points are valid the algorithm returns the unmodified path (Line 11). Otherwise} for each point $\posei$ in collision, we incrementally shift it towards safety according to \eqref{eq:update} until it is no longer in collision (Lines 4-6). Next, we interpolate between any shifted points and repeat the process for these new points (Lines 8-9). This is not guaranteed to be successful, so after a certain \rev{1.1, 3.3}{number of attempts, we revert to Fastron's repair step (Fig. \ref{fig:fastron-repair})for the non-validated portions of the path. Fastron's repair step takes a path, assuming the start and end are valid, and returns a repaired path that is guaranteed to be collision free \wrt traditional geometry-based collision checking. First, it uses traditional collision checking to find invalid path segments between valid point pairs. Next, it solves a series of small queries around invalid regions using RRT. Finally, the valid path is concatenated and returned. Our modification of the Faston repair step includes path's start and end points in the RRT search. The intuition behind the Fastron repair step is that queries between closer configuration points are easier to solve. While this is often true in practice, in general the smaller queries may be infeasible. To be able to find a solution if one exists, we perform a series of repair steps backing out through the portions of the path that are already certified valid (Lines 12-14). Finally, if the repair, which includes RRT query from the original start and end poses, fails to find a solution, we consider a query infeasible (Line 15).
}

\section{Results}

\begin{figure}[tb]
\centering
\begin{minipage}[t]{.66\textwidth}
	\begin{center}
	\begin{tabular}[t]{lr}
        \subfloat[Time v. batch to 150]{\includegraphics[width=0.485\linewidth,keepaspectratio=true]{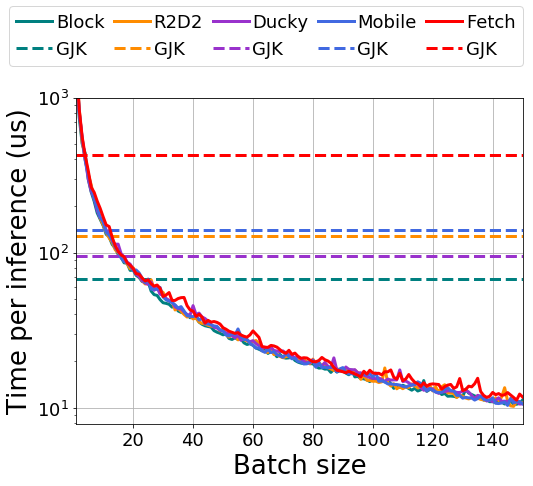}} &
        \subfloat[Time v. batch to 16K]{\includegraphics[width=0.485\linewidth,keepaspectratio=true]{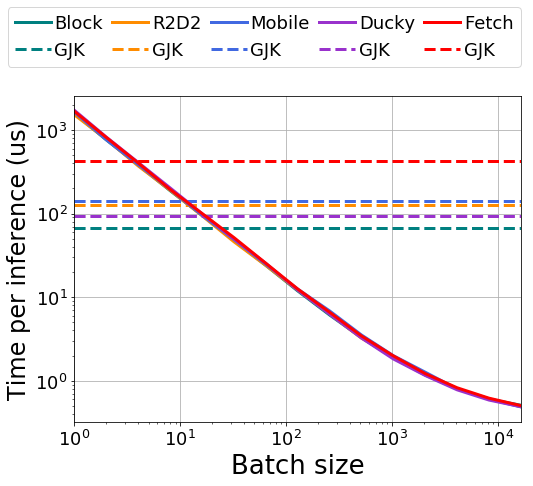}} \\
    \end{tabular}
    \caption{\footnotesize ClearanceNet inference time over batch size and model. Both y-axes and the second x-axis are logarithmic. Max speedups were 132, 261, 287, 193 and 845x respectively for Block, R2D2, Mobile, Ducky and Fetch.}
		\label{fig:batch_size}
	\end{center}
	
\end{minipage}\hfill
\begin{minipage}[t]{.3\textwidth}
  \includegraphics[width=1.0\linewidth]{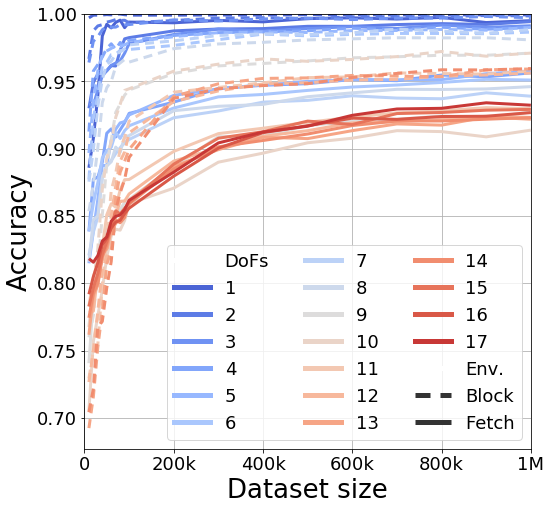}
  \caption{\footnotesize \rev{E.2, R1.2\ }{ClearanceNet accuracy over training set size and DoFs for Block and Fetch environments.}}
  \label{fig:dataset_vs_dofs}
\end{minipage}

\vspace{-1.5\baselineskip}
\end{figure}

\textbf{Setup:}
We evaluate ClearanceNet and CN-RRT on five environments (Fig. \ref{fig:envs}).
\textit{Block} and \textit{R2D2} contain two 7-DoF Kuka arms with fixed bases.
Block contains a floating object (Fig. \ref{fig:two_arms_block}), R2D2 a model of R2D2 (Fig. \ref{fig:two_arms_r2d2}). Planning DoFs and ClearanceNet input length are 14.
\textit{Ducky} contains a fixed-base 7-DoF Kuka arm and three objects (Fig. \ref{fig:ducky}),  selected to have different geometric properties, from the uniform block to the duck with smooth curved surfaces. The workspace is parameterized with the object locations. Thus planning is 7-DoF but the extended configuration space dimensionality (and ClearanceNet input length) is 16; 7 for the robot pose and 3 for each movable obstacle in the scene.
\textit{Mobile} consists of three mobile manipulators: Kuka arms attached to moving bases with 50 cm sides (Fig. \ref{fig:three_arms_mobile}). The collision checks between the three mobile manipulators are done as three pair-wise checks. Motion planning is 30-DoF but ClearanceNet input length is 17 for each pair-wise check; $7\times2$ for arm joints, one for base separation and two for base orientations.
\textit{Fetch} contains a Fetch robot, two sets of shelves and cubes in a 2m$\times$2m square. The shelves can slide 
and the cubes can be at any position on the floor. Motion planning is thus 11-DoF  
and ClearanceNet input length is 17. 


\subsection{ClearanceNet Results}
\textbf{Training:}
We generate datasets of $10^6$ training and $10^4$ evaluation samples for each environment using Bullet Physics Simulation \cite{pybullet}, which computes separation distances using the GJK distance algorithm~\cite{gjk1988}. Data collection takes 17-30 minutes per environment. The neural networks have layers with 1400 neurons. 

\textit{Hyperparameters:} 
We use Vizier \cite{vizier} with Gaussian Process Bandits \cite{gp-bandits} to search for learning hyperparameters using a total population of 1000 models.
A single network trains in one hour, while the hyperparameter search takes 24 hours on average. 
One set of optimized hyperparameters performs well across the environments \rev{3.2}{and robots}; thus we perform the tuning only once: dropout rate = 1\%, learning rate = 1.7495x$10^{-4}$, batch size = 191. \rev{3.2}{The overhead for applying CN-RRT to new robots is less than two hours, for most of which the robot can continue to operate as the model trains offline.}

\textit{Inference Speed:}
Time to perform a single collision check with ClearnanceNet decreases with batch size (Fig. \ref{fig:batch_size}), which we use to our advantage when building batched RRTs. It is interesting to note that, unlike classic methods, the network's size and therefore inference speed do not depend on the robot and workspace, suggesting that the method might be especially appropriate for environments with complex geometries and high-dimensional C-Spaces. 

\textit{Accuracy:}
\label{sec:learning_eval}
\rev{E2, R1.2}{Fig. \ref{fig:dataset_vs_dofs} shows ClearanceNet's accuracy for training set sizes up to $10^6$, with various DoFs frozen, for the most complicated (Fetch) and least complicated (Block) environments. This shows the required number of samples depends on both degrees of freedom and workspace complexity.}
Fig. \ref{fig:preds} displays ClearanceNet's accuracy with $10^6$ samples and summarizes classifications for $d^*=0.$ False negatives can lead to trajectories with collisions being classified as free, requiring more repair time. False positives can lead to conservative planners that find no paths at all; the adaptive threshold addresses this. 

\subsection{CN-RRT Results}
\textbf{Setup:} We compare CN-RRT to RRT \cite{rrts} with GJK collision checking \cite{gjk1988} (GJK-RRT), and RRT with Fastron~\cite{das2020fastron} collision checking (Fastron-RRT). \rev{1.4}{GJK is a geometry-based clearance finding algorithm, while Fastron is a learned collision checker that uses a support vector machine (kernel perceptron). Fastron-RRT includes a repair step that falls back on GJK-RRT to validate and correct paths.}

ClearanceNet is implemented with TensorFlow using the Python interface. Bullet is implemented in C++, and we use the Python interface PyBullet. Fastron is implemented in Python using Numpy. All evaluations run on an Nvidia Tesla V100 GPU with 16 Gbs of RAM, with fixed start and goal configurations averaged over 100 problems. \rev{R2.1}{Thresholds and threshold switch times are determined per-environment using Vizier \cite{vizier} with Gaussian Process Bandits \cite{gp-bandits}.}

\begin{table*}[tb]
\caption{\footnotesize Path characteristics and path-building statistics for CN-RRT and baselines averaged over 500 queries. CN-RRT-NG is an ablated version of CN-RRT with no gradient repair. \rev{R2.2}{Succ. Rate is the method's success at finding a path within the time limit.} \rev{R1.6}{Collision Checks are geometric checks, whereas Col. Checks Heuristic are checks by ClearanceNet and Fastron.} The $\%\Delta$ columns show percent increase or decrease as compared to GJK-RRT. The Path Length column group considers only successful queries, which biases it in favor of methods that solve only the shorter, easier problems.\label{tab:results}}

\scriptsize
\centering

\begin{tabular*}{\textwidth}{@{\extracolsep{\fill}}c|c|r|rrr|rrr|rr|rr}
\best{Env.} &  \best{Method} & \multicolumn{1}{c|}{\best{Succ.}} & \multicolumn{3}{c|}{\best{Compute Time}} & \multicolumn{3}{c|}{\best{Path Length}} &  \multicolumn{2}{c|}{\best{Collision}} & \multicolumn{2}{c}{\best{Col. Checks}} \\
 &   & \multicolumn{1}{c|}{\best{Rate}} & \multicolumn{3}{c|}{\best{(s)}} & \multicolumn{3}{c|}{\best{(steps)}} &  \multicolumn{2}{c|}{\best{Checks}} & \multicolumn{2}{c}{\best{Heuristic}}\\
 &        &  \multicolumn{1}{c|}{\best{(\%)}} & $\mu$ & $\%\Delta$ & $\sigma$ &    $\mu$ & $\%\Delta$ & $\sigma$ & $\mu$ & $\sigma$ &  $\mu$ & $\sigma$ \\\hline
 & CN-RRT & \best{97.0} & \best{2.0} & -27 & \best{5.2} & \best{964} & -21 & \best{398} & \best{7.4k} & \best{41.9k} & \best{19.0k} & 85.9k \\ 
\best{Block} & CN-RRT-NG & 95.6 & 2.4 & -14 & 6.3 & 1021 & -16 & 413 & 14.8k & 56.4k & 20.7k & 121.0k \\ 
(Fig. \ref{fig:two_arms_block}) & Fastron-RRT & 52.0 & 15.7 & +475 & 13.9 & 1169 & -4 & 484 & 11.8k & 48.7k & 43.6k & \best{43.4k} \\ 
& GJK-RRT & 95.2 & 2.7 & - & 6.7 & 1214 & - & 502 & 24.5k & 67.8k & - & - \\ 
\hline
 & CN-RRT & \best{98.8} & \best{1.6} & -42 & \best{3.5} & \best{926} & -26 & \best{414} & \best{0.5k} & \best{7.2k} & 39.6k & 122.2k \\ 
\best{R2D2} & CN-RRT-NG & 98.2 & 2.0 & -31 & 4.5 & 976 & -22 & 442 & 5.4k & 22.9k & 43.1k & 171.3k \\ 
(Fig. \ref{fig:two_arms_r2d2}) & Fastron-RRT & 97.6 & 3.4 & +19 & 5.0 & 1170 & -7 & 497 & 12.9k & 34.3k & \best{3.8k} & \best{3.4k} \\ 
 & GJK-RRT & 97.6 & 2.8 & - & 5.2 & 1259 & - & 523 & 14.5k & 31.0k & - & - \\ 
\hline
 & CN-RRT & \best{91.0} & \best{4.3} & -12 & 8.7 & \best{673} & -36 & 336 & 18.8k & 52.2k & \best{6.8k} & 14.0k \\ 
\best{Ducky} & CN-RRT-NG & 90.4 & 4.8 & -2 & 9.0 & 941 & -11 & 428 & 26.7k & 59.2k & 6.8k & 11.3k \\ 
(Fig. \ref{fig:ducky}) & Fastron-RRT & 1.0 & 31.3 & +545 & \best{2.8} & 741 & -30 & \best{309} & \best{0.1k} & \best{0.7k} & 47.4k & \best{5.2k} \\ 
 & GJK-RRT & 89.4 & 4.9 & - & 9.2 & 1055 & - & 433 & 25.7k & 51.3k & - & - \\ 
\hline
 & CN-RRT-NG & \best{70.7} & 91.6 & +2 & 67.3 & \best{3799} & -10 & \best{1824} & 306.2k & 265.4k & \best{295.4k} & 524.6k \\ 
\best{Mobile} & Fastron-RRT & 0.0 & 180.4 & +101 & \best{0.3} & - & - & - & \best{0.0k} & \best{-} & 481.4k & \best{83.3k} \\
(Fig. \ref{fig:three_arms_mobile}) & GJK-RRT & 63.5 & \best{89.8} & - & 73.4 & 4205 & - & 2099 & 411.1k & 338.0k & - & - \\ 
\hline
 & CN-RRT-NG & \best{70.0} & \best{416.8} & -30 & 417.5 & 2857 & +12 & 1454 & 436.7k & 435.6k & \best{120.7k} & \best{140.1k} \\ 
\best{Fetch} & Fastron-RRT & 8.0 & 923.1 & +54 & \best{266.7} & \best{1324} & -48 & \best{781} & \best{24.8k} & \best{163.8k} & 1268.2k & 423.6k \\ 
(Fig. \ref{fig:fetch}) & GJK-RRT & 52.0 & 599.2 & - & 426.4 & 2553 & - & 1329 & 617.5k & 439.3k & - & - \\ 
\hline

\end{tabular*}
\vspace{-2.5\baselineskip}
\end{table*}

\begin{figure}[p]
\vspace{-1.2\baselineskip}
\begin{minipage}[t]{.32\textwidth}
\vspace{0pt}
\newcommand\predshgt{2.5cm}
    \begin{tabular}{c}
    \subfloat[\scriptsize Block, 95.5\% Input size: 14]{\makebox[\textwidth][c]{\includegraphics[height=\predshgt]{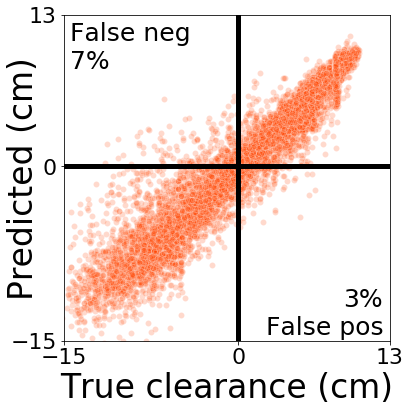}}}
    \label{fig:floating_block_preds} \\
    
    \subfloat[\scriptsize R2D2, 96.0\% Input size: 14]{\makebox[\textwidth][c]{\includegraphics[height=\predshgt]{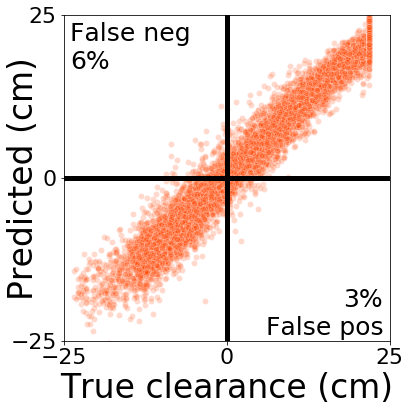}}}
    \label{fig:r2d2_preds} \\
    
    \subfloat[\scriptsize Ducky, 91.4\% Input size: 16]{\makebox[\textwidth][c]{\includegraphics[height=\predshgt]{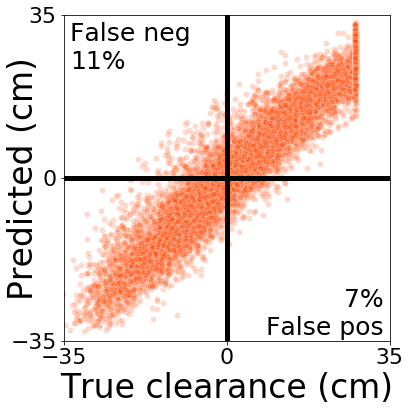}}}
    \label{fig:ducky_preds} \\
    
    \subfloat[\scriptsize Mobile, 96.3\% Input size: 17]{\makebox[\textwidth][c]{\includegraphics[height=\predshgt]{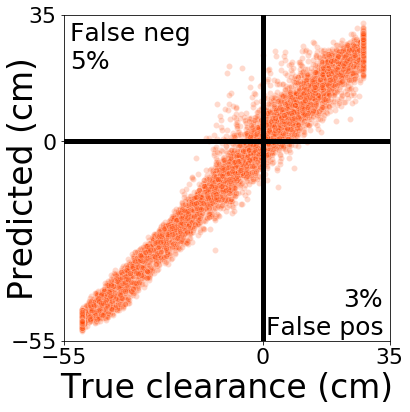}}}
    \label{fig:mobile_base_preds} \\
    
    \subfloat[\scriptsize Fetch, 95.1\% Input size: 11]{\makebox[\textwidth][c]{\includegraphics[height=\predshgt]{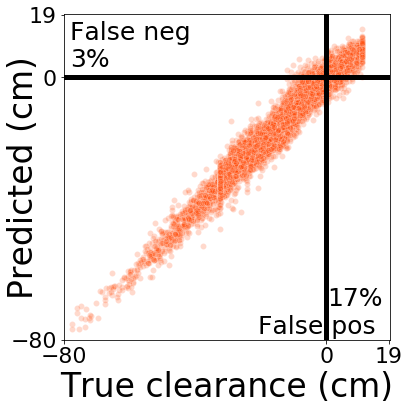}}}
    \label{fig:fetch_preds} \\
\end{tabular}
	\caption{\footnotesize ClearanceNet accuracy. Quadrants are true positive collision (bottom left), true negative collision-free (top right), false positive (bottom right), and false negative (top left). 
	\label{fig:preds}}

\end{minipage}\hfill
\begin{minipage}[t]{.66\textwidth}
\vspace{0pt}
\newcommand\hgt{2.5cm}
\begin{center}
	\begin{tabular}{cc}
  \subfloat[\scriptsize Block success rate]{\includegraphics[height=\hgt]{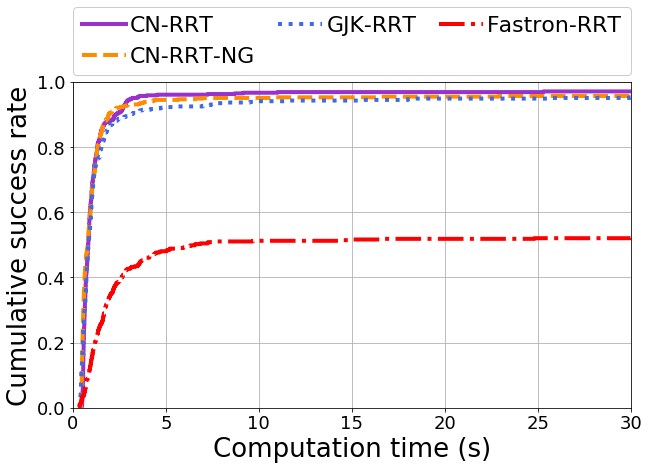}} &
  \subfloat[\scriptsize Block time allocation]{\includegraphics[height=\hgt]{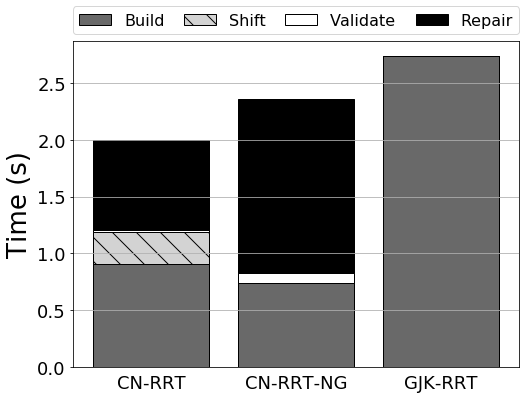}} \\
  \subfloat[\scriptsize R2D2 success rate]{\includegraphics[height=\hgt]{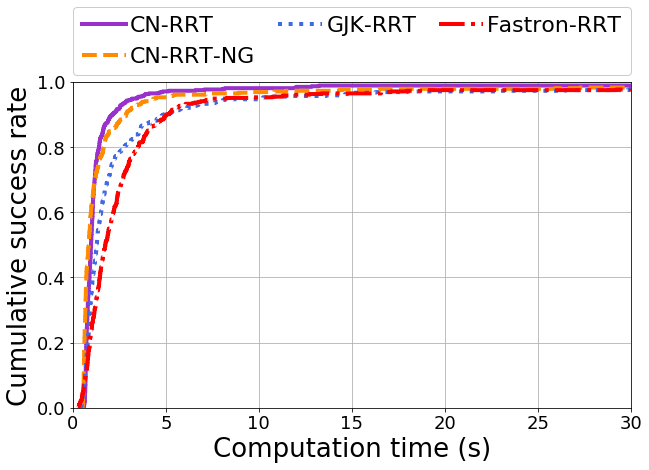}} &
  \subfloat[\scriptsize R2D2 time allocation]{\includegraphics[height=\hgt]{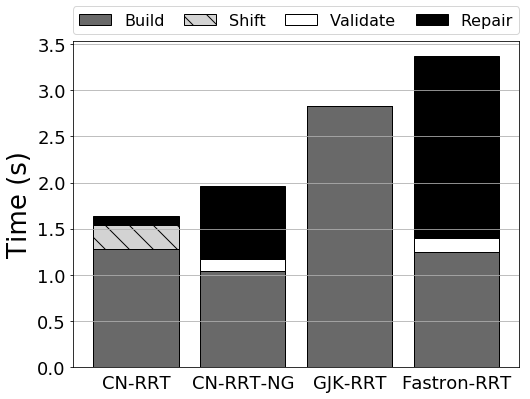}} \\
  \subfloat[\scriptsize Ducky success rate]{\includegraphics[height=\hgt]{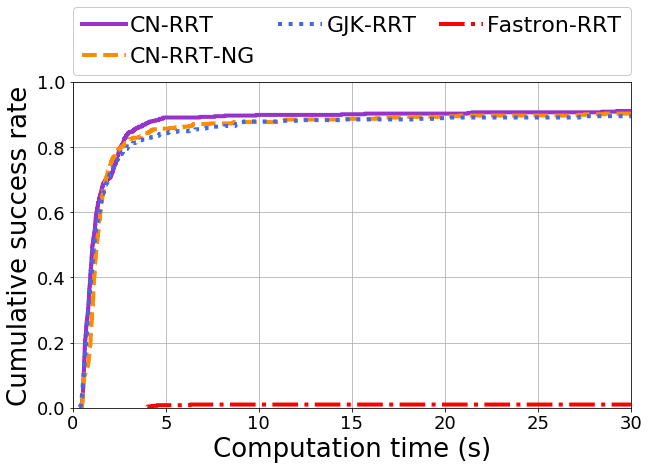}} &
  \subfloat[\scriptsize Ducky time allocation]{\includegraphics[height=\hgt]{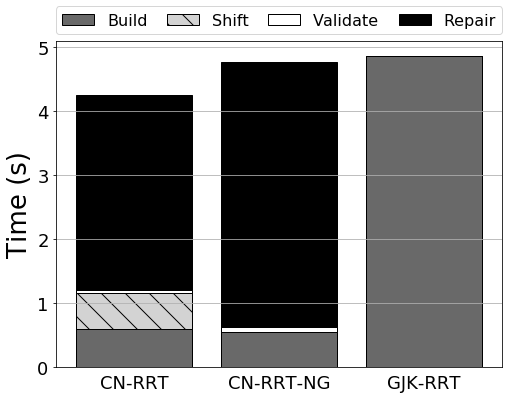}} \\
  \subfloat[\scriptsize Mobile success rate]{\includegraphics[height=\hgt]{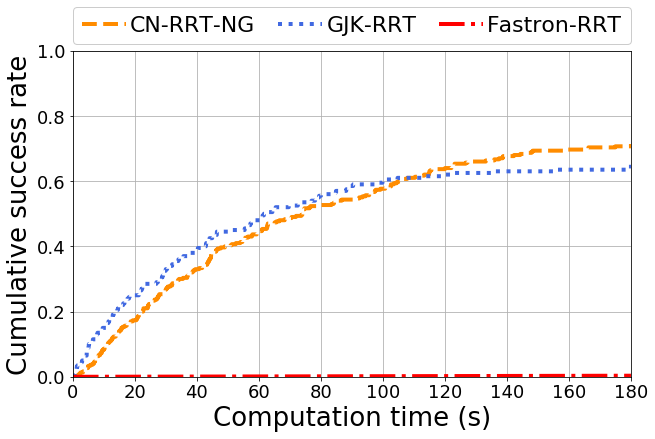}} &
  \subfloat[\scriptsize Mobile time allocation]{\includegraphics[height=\hgt]{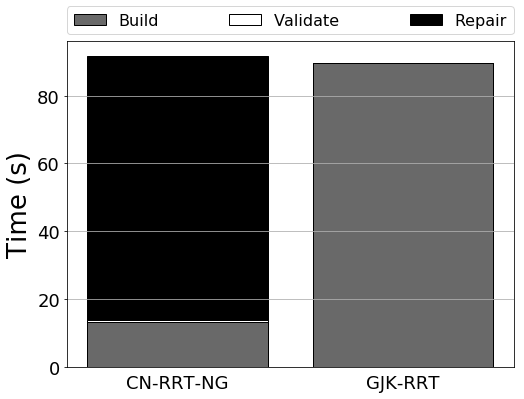}} \\
  \subfloat[\scriptsize Fetch success rate]{\includegraphics[height=\hgt]{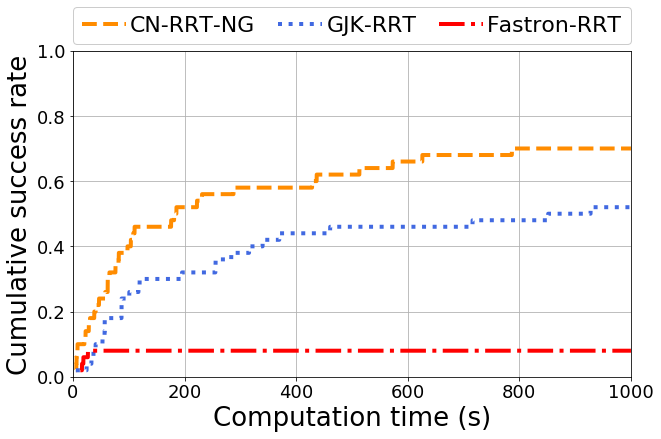}} &
  \subfloat[\scriptsize Fetch time allocation]{\includegraphics[height=\hgt]{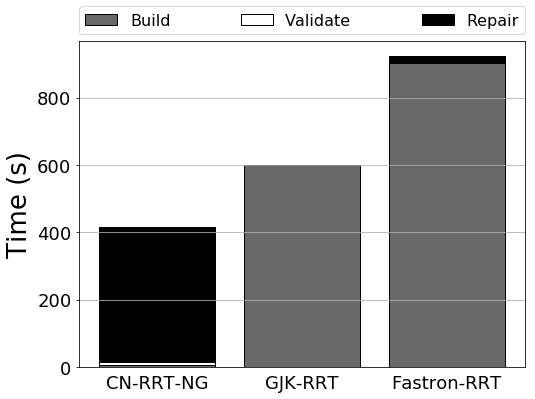}} \\
  
    \end{tabular}
\caption{Success rate over time and time per phase. The bar graphs show average time spent in each phase: \rev{R1.5}{constructing the initial RRT (Build), repairing the path with the network gradient (Shift), checking the path with GJK (Validate) and replanning broken segments using GJK-RRT (Repair).} Fastron is included when its scale is comparable to the other methods. 
\label{fig:rrt}}

\end{center}
\end{minipage}
\end{figure}

\textbf{Success rate and planning speed:} 
CN-RRT is more likely to find a solution for a query than the comparison methods most of the time (Fig. \ref{fig:rrt}). In the Ducky environment the performance is comparable to GJK. In the Mobile environment, which contains more robots in the evaluation than during training, GJK solves more easier solutions faster. However, as the queries become more difficult CN-RRT finds solutions that the comparison methods could not. In all other environments CN-RRT consistently solves more queries faster (Table \ref{tab:results}). 

\textbf{Time allocation:}
In all trials, failure to solve a problem counts as the maximum time allotted. CN-RRT solves a single query faster than the baselines for all environments except Mobile (Fig. \ref{fig:rrt}). Additionally, the repair time for CN-RRT is shorter than the build time for GJK-RRT on average, demonstrating that repairing a path is faster than building one from scratch.

\textbf{Path length:}
Next, we examine the quality of the paths that CN-RRT finds. In all environments except Fetch, CN-RRT paths are shorter on average than GJK-RRT and Fastron-RRT (Table \ref{tab:results}). Although we do not explicitly optimize for path length, ClearanceNet is biased toward higher-clearance paths, which are likely to be shorter in relatively uncluttered environments.
In the more cluttered Fetch environment, GJK-RRT and Fastron-RRT have shorter paths on average, although this is affected by the fact that both algorithms solved only an easier subset of the problems CN-RRT solved.

\textbf{Gradient-based repair:}
There are two parameters we can tune: step size $\alpha$ and extra steps $\nextra$. We run a grid of experiments per-environment on a training set of motion planning problems to determine the best values for these parameters. For Block $\alpha = 0.15$ and $\nextra = 0$. For R2D2 and Ducky, $\alpha = 0.05$ and $\nextra = 3$. 
Gradient-based repair is inapplicable to Mobile because the collision checks are done pair-wise, and to Fetch because the robot is non-holonomic (the base cannot translate sideways). For these environments we use the ablation method CN-RRT-NG.
For all other environments, gradient-based repair is an improvement over GJK-based repair in terms of both average calculation time and average path duration (Fig. \ref{fig:rrt} CN-RRT-NG).

\textbf{On robot experiments:}
We validate CN-RRT by taking paths it finds and executing them on a real Fetch robot. The start poses are randomly generated, and the goal poses are selected to be especially challenging: they require reaching into enclosed shelves (Fig. \ref{fig:fetch_real}). For one randomly selected query, CN-RRT's planning time is \dur{47}, while the comparison methods fail to find a solution.

\section{Analysis and Discussion}
\textbf{Probabilistic completeness:}
Algorithm \ref{alg:parallelConnect} is probabilistically complete. We can consider three cases for the $\tcnn:$ 1) the classifier classifies all points as in collision, 2) all points as out of collision, and 3) $\tcnn$ falls somewhere in the middle. In the first case, CN-RRT will time out and proceed with \rev{1.1}{its repair loop (Algorithm \ref{alg:repair} Lines 12-14)} acting from $\startr$ to $\goalr.$ In the second case, the algorithm reduces to a lazy RRT and again falls back on the repair \rev{1.1}{loop. In the last case, Algorithm \ref{alg:parallelConnect} cheaply finds a candidate path which is passed to Algorithm \ref{alg:repair} for verification and possible repair.
The invariant for the while loop (Algorithm \ref{alg:repair} Lines 2-10) is that all points in path $\epath$ with indices $[0,..,i-1]$ are certified valid with GJK. Upon exit, the algorithm either returns because the whole path is certified valid, or it enters the repair loop. The invariant of the repair loop is that $\epath[0, j]$ is certified valid and $\epath[j, \epath.length]$ is invalid. Our modification of Fastron repair performs RRT with geometry-based collision checking on $\epath[j, \epath.length].$ Therefore if the repair step of Line 13 is successful, it returns path $\epath[0, j]$ and $\epath[j, \epath.length]$ is valid. Otherwise, probabilistically complete RRT found no path between $\epath[0, \epath.length],$ and we return that the query was infeasible on Line 15 in Algorithm \ref{alg:repair}. 
Thus, Algorithm \ref{alg:parallelConnect} is probabilistically complete and returns a valid path iff RRT with geometry-based collision checking finds a path.}

\textbf{Neural networks as heuristics:}
Separation distance is a Lipschitz continuous function when the robots and obstacles are rigid bodies \cite{Lavalle06book}. CN-RRT exemplifies several benefits of choosing a neural network as a function approximation of a Lipschitz continuous function. First, neural networks running on GPUs are exceptionally efficient at processing large batches of data, and conveniently motion planning is well-suited for batched processing \cite{nancy-parallel}. Second, neural networks are differentiable functions, and thus the decision making manifold (contact surface defined with threshold $d^*$ selection) is ``well-behaved'' (Lipschitz continuous and monotonic), allowing a simple and effective algorithm conservativeness tuning. Last, the differentiability of the approximator comes in handy for recommending incremental path improvements in the direction of its gradient. Future work can look into approximating other complex-to-compute Lipschitz continuous measures in motion planning and exploit the same benefits (large data batching, decision-making manifold adaptation and gradient-based improvements) to make motion planning more efficient. 

\textbf{Run-time trade-off analysis:} 
The biggest downside of neural networks as approximators is the one-time cost of \rev{1.3}{data collection and} training, which may be justified if they are used in enough queries. We explore the trade-offs between cost of training, collision checking speed up and number of collision checks, and answer how many queries are needed to offset the cost of training. \rev{1.3}{Training and data collection can be, and in practice often are, done offline in a process concurrent to robot operations, without blocking a robot. Thus, the number of queries $\numqueries$ needed to offset neural network overhead is
$\numqueries > \frac{\tinvest}{\Delta t} + \frac{\toffline}{\trrt}, \;\; \Delta t = \trrt - \tcnrrt$, where $\tinvest$, $\toffline$ are online and offline time investment and $\trrt$, $\tcnrrt$ are expected query completion time for RRT and CN-RRT.}

\begin{figure}[]
\centering
  \includegraphics[width=1.0\linewidth]{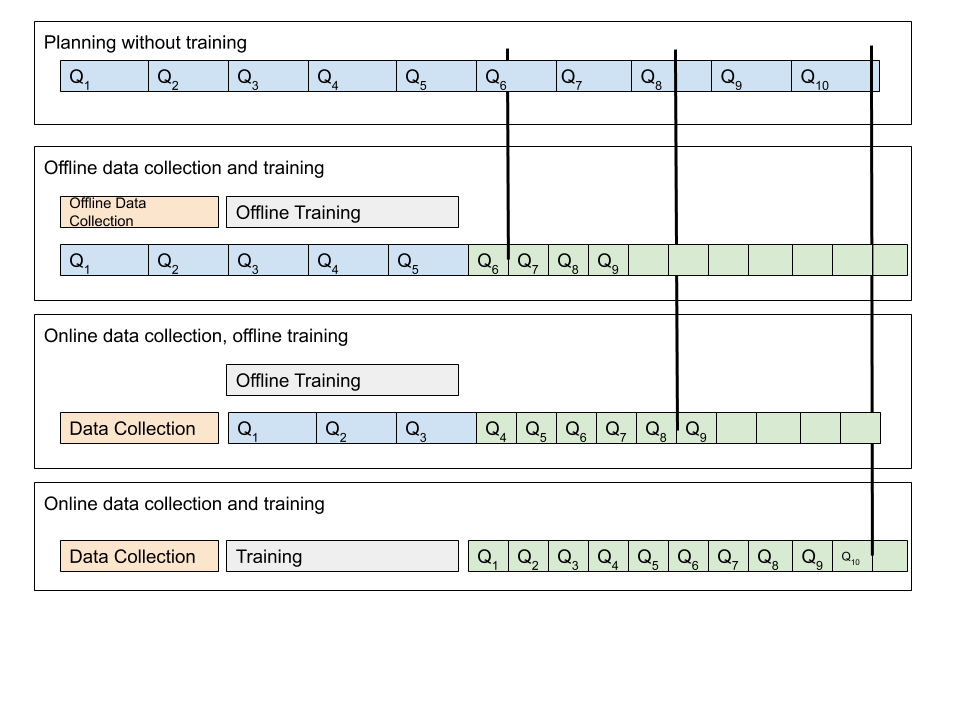}
  \caption{\footnotesize Illustration of queries needed to justify training investment for online data collection and training (bottom), online data collection and offline training (middle), and offline data collection and training (top) compared with non-learned classic planning. Blue blocks represent duration of classic motion planning queries, green CN-RRT queries, grey training, and orange data collection. Left to right is time, and parallel tracks represent concurrent processes.}
  \label{fig:tradeoff}
\vspace{-2\baselineskip}
\end{figure}

The cost of motion planning with RRTs is $\trrt \sim \numcd * \tgjk,$ where $\numcd =$ expected number of collision checks and $\tgjk =$ time for a single geometric collision check. The time to solve one CN-RRT query is $\tcnrrt \sim \numcd * \tgjk * \ffp * \fspeedup + \numcd * \tgjk * \ffn * \pathlength,$ where: $\ffp \geq 1$ is the increase factor in collision checks due to $\tcnn$ mistakenly classifying non-collisions as collisions (i.e. the cost of false positives); $\ffn \geq 1$ is the increase in collision checks due to points being misclassified as free, triggering repair (i.e. the cost of false negatives); $\pathlength << 1$ is the ratio of expected path length to total number of collision checks; $\fspeedup$ is computation speed up due to batching, $0 < \fspeedup << 1,$ as a ratio between time to compute a single collision check with a neural network (in a batch) and $\tgjk$. 
After substituting the above and arranging the terms, we arrive at:
\begin{equation}
    \label{eq:tradeoff}
    \numqueries > \frac{\tinvest / \tgjk}{\numcd ( 1 - \ffp \fspeedup  - \ffn \pathlength)} + \frac{\toffline / \tgjk}{\numcd}
\end{equation}
The cost of training consists of time for data collection ($\datasetsize \tgjk$) and time for training ($\ftraining \tgjk$), where $\datasetsize =$ dataset size and $\ftraining > 1$ is the training time expressed as a multiple of $\tgjk.$ We consider two options:
a) data collection is online and training is offline ($\tinvest = \datasetsize \tgjk$ is time for data collection, and $\toffline = \ftraining \tgjk$ is time for training), and b) both are online ($\toffline = 0,\, \tinvest = (\datasetsize + \ftraining) \tgjk$). 
See Fig. \ref{fig:tradeoff} for an illustration.
Eq. \eqref{eq:tradeoff} yields the following conclusions:
\begin{enumerate}
    \item \textit{Precision/recall trade off:} The classifier's false negative rate is more important, since $\pathlength >> \fspeedup,$ justifying the adaptive thresholding in Alg. \ref{alg:parallelConnect}. Here, $\pathlength$ in $10^{-2} - 10^{-1}$ range (Table \ref{tab:results}), while $\fspeedup \sim 10^{-3}$ (Fig. \ref{fig:batch_size}).  
    \item \textit{Training trade-off:} There is a linear dependency between training time and data collection. Thus, it is worthwhile to look for more sample-efficient and faster training methods. In this paper, data collection requires about 30 minutes, and training about 60 minutes for a total of 90 minutes of overhead.
    \item \textit{Motion planing complexity:} CN-RRT fares better at complex queries that require many collision checks, because $\numqueries$ is inversely proportional to $\numcd.$ 
    \item \textit{Knowing $\numqueries$} allows us to tailor the training. For small $\numqueries$, we can use cheap training and inaccurate models (increasing $\ffp$ and $\ffn$); for large $\numqueries$, we can invest in longer, more precise training. This suggests future research in active sampling and interactive training. For our environments, \rev{1.3}{we estimate $\trrt, \tgjk, \Delta t, \datasetsize$ from Table \ref{tab:results} and Fig. \ref{fig:dataset_vs_dofs} and derive that in the the fully online setting $\numqueries$ is in the thousands, with the exceptions of Fetch, which requires fewer than 25 queries, and Mobile, which provides no savings. In the more realistic, offline training case, the required number of queries drops to 5 for Fetch, and less than a thousand for the simpler environments. See Appendix for the full analysis.}
\end{enumerate}

\textbf{Sampling vs. inference complexity:} Interestingly, $\fspeedup$ depends only on DoFs and not on the geometric properties of workspace objects. The topological complexities of the configuration space are addressed through the training set. For a good separation distance approximator, it is imperative that the training samples are from relevant and interesting regions. This suggests future work applying different sampling heuristics to collect high-quality training sets. Here we use uniform random sampling, but more appropriate sampling would likely result in better models, or require fewer samples to achieve the same accuracy.   

\textbf{Limitations:} 
The two main features missing from CN-RRT are adaptation to moving obstacles and adaptation to new workspace objects. CN-RRT can easily be adapted for moving obstacles; the only change required is to let the workspace configuration $\posew$ change during motion planning in Algorithm \ref{alg:parallelConnect}. The second is less straightforward. One avenue to address this is to use domain adaptation and learn to extend the network for additional degrees of freedom and perform additional training. Another approach is to use sensor input, which provides up-to-date information on the environment, but may also be high-dimensional. 

\rev{3.1,3.2}{\textbf{Future work:} An interesting extension would be to implement a lifelong planner that starts as a classic planner, then over its lifetime trains a collision checking model with data collected during planning, or generalizes between different robots of similar geometries.} \rev{R 3.5}{A different direction could make the gradient repair step in a specific direction to target specific robot joints.}
\section{Conclusions}
%
We presented ClearanceNet, a neural network approximator for separation distance in extended configuration space, and CN-RRT, an RRT algorithm that exploits efficient batch processing and neural network differentiability. CN-RRT adds multiple edges at a time, adaptively relaxes clearance requirements for difficult queries, and uses ClearanceNet's gradient for repairs. Evaluated on five environments, CN-RRT produces shorter paths more quickly. On-robot experiments demonstrate its applicability to real robots. Finally, we analyzed the algorithm's trade-offs and offered ideas for future work using neural networks as approximators for Lipschitz continuous measures in motion planning. 

\section{Acknowledgements}
The authors thank H.T.L. Chiang for helpful discussions, L. Downs and K. Reymann for asset creation, and anonymous reviewers for thoughful comments.

\bibliographystyle{abbrv}
\bibliography{references}  

\section{Appendix}

\subsection{Analysis of number of queries needed to justify offline investment}

In reality, neural network training happens offline, and therefore the robot downtime is only to collect the dataset. Applying Eq. (\ref{eq:tradeoff}) directly is difficult because of the unknown terms for $f_{\text{fp}}$ and $f_{\text{fn}}$ penalties. Instead, we estimate the numbers starting from 
\begin{equation}
    \numqueries > \frac{\tinvest}{\Delta t} + \frac{\toffline}{\trrt}, \;\; \Delta t = \trrt - \tcnrrt
    \label{eq:1}
\end{equation}
using the empirical data. We estimate the number of queries in three different ways. 

\begin{enumerate}
    \item \textit{Online data collection and offline training:} $\tinvest = \datasetsize \tgjk$ and $\toffline = \ftraining \tgjk$ is time for training
    \begin{equation}
    \numqueries > \frac{\datasetsize \tgjk}{\Delta t} + \frac{t_{\text{training}}}{\trrt}
     \label{eq:2}
\end{equation}
    \item \textit{Both data collection and training are online:} $\toffline = 0$ and $\tinvest = (\datasetsize \tgjk + t_{\text{training}}$.
    \begin{equation}
    \numqueries > \frac{\datasetsize \tgjk + t_{\text{training}}}{\Delta t}.
     \label{eq:3}
\end{equation}
\item \textit{Both data collection and training are offline:} $\toffline = \datasetsize \tgjk + t_{\text{training}}$ and $\tinvest = 0$.
    \begin{equation}
    \numqueries > \frac{\datasetsize \tgjk + t_{\text{training}}}{\trrt}.
     \label{eq:4}
\end{equation}
\end{enumerate}
We use the evaluation results to estimate. 
\begin{itemize}
    \item $t_{\text{training}}$ empirically measured.
    \item $\datasetsize$ from Fig. \ref{fig:batch_size} and uniform one million.
    \item $\tgjk$ from Fig. \ref{fig:dataset_vs_dofs}.
    \item $\Delta t$ from Table \ref{tab:results}, as a difference between GJK-RRT and CN-RRT in column Compute Time.
    \item $\trrt$ from Table \ref{tab:results}, column Compute Time.
\end{itemize}

\begin{table*}[h]
\caption{\footnotesize Results of analysis of queries needed to justify offline investment.\label{tab:analysis}}

\scriptsize
\centering
\begin{tabular}{l|r|r|r|r|r|r|>{\bfseries}r|>{\bfseries}r|>{\bfseries}r|>{\bfseries}r}
Env	& $t_{\text{training}}$ & $\datasetsize$ & $\tgjk$ & $\trrt$ & $\tcnrrt$ & $\Delta t$ & 	Eq. (\ref{eq:2}) & Eq. (\ref{eq:3}) &	Eq. (\ref{eq:4}) \\
  & (s) & (x$10^3\,$s) & (x$10^{-3}\,$s) & (s) &  &  (s) &   &  &    \\\hline
Block	&1500&	350&	0.1&	2.70&	2.00 &	0.7&	606&	2,193&	569	\\
R2D2	&1800&	400&	0.2&	2.80&	1.60.00&	1.2&	710&	1,567&	671	\\
Ducky	&1800&	400&	0.2&	4.90&	4.30&	0.6&	501&	3,133&	384	\\
Mobile	&1800&	500&	0.2&	89.90&	91.70&	-1.8&	-36&	-1,056&	21	\\
Fetch	&2400&	600&	1.0&	599.20&	417.20&	182&	7&	16&	5	\\\hline
Block	&3600&	1000&	0.1&	2.70&	2.00&	0.7&	1,476&	5,286&	1,370	\\
R2D2	&3600&	1000&	0.2&	2.80&	1.60&	1.2&	1,452&	3,167&	1,357	\\
Ducky	&3600&	1000&	0.2&	4.90&	4.30&	0.6&	1,068&	6,333&	776	\\
Mobile	&3600&	1000&	0.2&	89.90&	91.70&	-1.8&	-71&	-2,111&	-42	\\
Fetch	&3600&	1000&	1.0&	599.20&	417.20&	182&	12&	25&	8	\\
\end{tabular}
\vspace{-2.5\baselineskip}
\end{table*}

Table \ref{tab:analysis} summarizes the results of the analysis for the reasonable choices of dataset (derived from Fig. \ref{fig:dataset_vs_dofs}) and the conservative dataset of one million samples we used in the evaluations.

In all cases, including the large dataset of million samples, we need to perform fewer than seven queries to collect the dataset. This is because RRT performs a large number of collision checks. 
\begin{enumerate}
    \item \textit{Online data collection and offline training (Eq. (\ref{eq:2})):} This is the most realistic case. The easy environments such as Block, R2D2, and Ducky, in which collision checks and query computation are relatively fast, require up to 1500 queries for the large dataset, and up to approximately 700 queries for the optimized dataset size. However, the most complex environment, Fetch, requires less than 15 queries. In Mobile environment, CN-RRT is slower than RRT, and although it produces shorter paths, it yields no time savings.  
    \item \textit{Both data collection and training are online (Eq. (\ref{eq:3})):} This is the most pessimistic use case, and it yields results similar to the offline training use case above with a multiplicative factor. Fetch environment still requires fewer than 25 queries, while the simpler environments require several thousands. 
    \item \textit{Both data collection and training are offline (Eq. (\ref{eq:4})):} This use case is most optimistic, but in practice it might require a different strategy for data collection. We provide the analysis here for completeness and to motivate future work. In this case, all environments see time savings except Mobile. The number of queries is less than 1500 even in the most pessimistic cases, and Fetch requires fewer than eight.
\end{enumerate}

Overall, the data collection is not as expensive as it appears at first glance, and for the more complex environments CN-RRT requires only a modest number of queries to justify offline investment.

\subsection{Fastron repair diagram}

\end{document}